%% file: acl_latex.tex
\pdfoutput=1

\documentclass[11pt]{article}
\usepackage{acl}
\usepackage{times}
\usepackage{latexsym}

\usepackage[T1]{fontenc}
\usepackage[utf8]{inputenc}
\usepackage{microtype}

\usepackage{booktabs}
\usepackage{color}
\usepackage{colortbl}

\usepackage{graphicx}
\usepackage{amssymb}
\usepackage{enumitem}

\title{What Makes Reading Comprehension Questions Difficult?}

\author{Saku Sugawara,$^1$ Nikita Nangia,$^2$ Alex Warstadt,$^2$ Samuel R. Bowman$^2$ \\
  $^1$National Institute of Informatics, $^2$New York University \\
  \texttt{saku@nii.ac.jp, \{nikitanangia,warstadt,bowman\}@nyu.edu}}

\newcommand{\checkedbox}{$\rlap{$\checkmark$}\square$~}
\newcommand{\checkbox}{$\rlap{\hphantom{$\checkmark$}}\square$~}

\date{}

\begin{document}
\maketitle
\begin{abstract}
For a natural language understanding benchmark to be useful in research, it has to consist of examples that are diverse and difficult enough to discriminate among current and near-future state-of-the-art systems.
However, we do not yet know how best to select text sources to collect a variety of challenging examples.
In this study, we crowdsource multiple-choice reading comprehension questions for passages taken from seven qualitatively distinct sources, analyzing what attributes of passages contribute to the difficulty and question types of the collected examples.
To our surprise, we find that passage source, length, and readability measures do not significantly affect question difficulty.
Through our manual annotation of seven reasoning types, we observe several trends between passage sources and reasoning types, e.g., logical \mbox{reasoning} is more often required in questions written for technical passages.
These results suggest that when creating a new benchmark dataset, selecting a diverse set of passages can help ensure a diverse range of question types, but that passage difficulty need not be a priority.
\end{abstract}

\section{Introduction}

State-of-the-art systems have shown performance comparable with humans on many recent natural language understanding (NLU) datasets \cite{devlin-etal-2019-bert,sun-etal-2021-ernie}, suggesting that these benchmarks will no longer be able to measure future progress.
To move beyond this, we will need to find better ways of building difficult datasets, ideally without sacrificing diversity or coverage \cite{bowman-dahl-2021-will}.
To obtain such human-written examples at scale, there are active lines of crowdsourcing research on protocols of worker handling and feedback \cite{nangia-etal-2021-ingredients} and the design of the collection task \cite{ning-etal-2020-torque,rogers-etal-2020-getting}.
However, we do not have clear information on what aspects of \textit{text sources} affect the difficulty and diversity of examples.

\begin{figure}[t]
    \centering \small
    \fbox{\parbox{0.95\linewidth}{
        \textbf{MCTest}: Tony walked home from school on his birthday. He was surprised to see a lot of cars in front of his house. When he opened the door and entered the house, he heard a lot of people yell, ``Surprise!'' It was a surprise party for his birthday.  His parents called all his friends' parents and invited them to come to a party for Tony. [...] \\  
        Q: \textit{Who were invited to the party and by who?} \\
        \checkbox \textit{Tony's parents invited only his friends} \\
        \checkbox \textit{Tony invited his friends and their parents} \\
        \checkbox \textit{Tony's parents invited his friends' parents} \\
        \checkedbox \textit{Tony's parents invited his friends and their parents}
    }}
    \fbox{\parbox{0.95\linewidth}{
        \textbf{ReClor}: Humanitarian considerations aside, sheer economics dictates that country X should institute, as country Y has done, a nationwide system of air and ground transportation for conveying seriously injured persons to specialized trauma centers. Timely access to the kind of medical care that only specialized centers can provide could save the lives of many people. [...] \\ 
        Q: \textit{What is the economic argument supporting the idea of \hphantom{Q:}a transportation system across the nation of Country X?} \\
        \checkbox \textit{Building the transportation system creates a substantial \hphantom{\checkbox}increase of jobs for the locals} \\
        \checkedbox \textit{Increasing access to specialized medical centers can \hphantom{\checkbox}lower the chance of the workforce population dying} \\
        \checkbox \textit{Transportation ticket prices directly contribute to the \hphantom{\checkbox}government's revenue} \\
        \checkbox \textit{Country Y was successful with their attempts to poten-\hphantom{\checkbox}tially save lives so Country X should try it as well}
    }}
    \caption{
        Example questions for passages from simple narratives (MCTest) and technical arguments (ReClor).
    }
    \label{fig:intro-example}
\end{figure}

Crowdsourced datasets in reading comprehension use passages taken from a variety of sources, such as news articles, exams, and blogs, about which questions are written \cite{lai-etal-2017-race,trischler-etal-2017-newsqa,rogers-etal-2020-getting}.
The first example in Figure~\ref{fig:intro-example} is from MCTest~\cite{richardson-etal-2013-mctest}, the passages of which are written in grade-school-level English.
The second example is from ReClor~\cite{Yu2020ReClor}, which consists of passages and questions written for graduate and law school admission examinations.
We hypothesize that difficult passages, such as those in the second example, are more suitable for crowdsourcing challenging questions.
Passages that are linguistically complex and have dense information could help facilitate the writing of questions that require understanding a wide range of linguistic and world knowledge, following intricate events, and comprehending logical arguments.
In contrast, easy passages, as in children's stories, likely talk about common situations and simple facts, which might prevent workers from writing difficult questions.

In this work, we crowdsource multiple-choice reading comprehension questions to analyze how question difficulty and type are affected by the choice of source passage.
Using passages extracted from seven different sources, we ask crowdworkers to write questions about the given passages.
We compute the difference between human and machine accuracy, using it as a measure of the question difficulty, to investigate whether there is a correlation between the question difficulty and linguistic aspects of the passage, such as their source, length, and readability.

In addition to a standard setting where we directly accept crowdworkers' submissions, we use an adversarial setting in which they have to write questions that fool a strong reading comprehension model \cite{bartolo-etal-2020-beat,kiela-etal-2021-dynabench}.
Previous work finds that questions that require numerical reasoning frequently appear in the adversarial data collection of the extractive QA task on Wikipedia articles \cite{kaushik-etal-2021-efficacy}, but our aim is to see whether we observe a similar trend in multiple-choice questions written for different passage sources or if the adversarial setting is useful for collecting especially diverse questions.

To our surprise, we find that the difficulty of collected questions does not depend on the differences of passages in linguistic aspects such as passage source, passage length, Flesch--Kincaid grade level \cite{kincaid1975derivation}, syntactic and lexical surprisal, elapsed time for answering, and the average word frequency in a passage.
Our main positive finding comes through our manual annotation of the types of reasoning that each question targets, where we observe that questions that require numerical reasoning and logical reasoning are relatively difficult.
In addition, we find several trends between the passage sources and reasoning types.
For example, logical reasoning is more often required in questions written for technical passages, whereas understanding of a given passage's gestalt and the author's attitude toward it are more frequently required for argumentative and subjective passages than expository passages.

These results suggest that when creating a new benchmark dataset or choosing one for evaluating NLU systems, selecting a diverse set of passages can help ensure a diverse range of question types, but that passage \textit{difficulty} need not be a priority.
Our collected datasets could be useful for training reading comprehension models and for further analysis of requisite knowledge and comprehension types in answering challenging multiple-choice questions.\footnote{Our datasets, annotation instructions and results, and crowdsourcing scripts are available at \url{https://github.com/nii-cl/qa-text-source-comparison}.}

\section{Related Work}

\paragraph{Crowdsourcing NLU Datasets}
Crowdsourcing has been widely used to collect human-written examples at scale \cite{rajpurkar-etal-2016-squad,trischler-etal-2017-newsqa}.
Crowdworkers are usually asked to write questions about a given text, sometimes with constraints imposed to obtain questions that require specific reasoning skills such as multi-hop reasoning \cite{yang-etal-2018-hotpotqa} or understanding of temporal order, coreference, or causality \cite{rogers-etal-2020-getting}.
In this study, to analyze naturally written examples, we do not consider specific constraints on questions or answer options. 

Current benchmark datasets constructed by crowdsourcing may not be of sufficient quality to precisely evaluate human-level NLU.
For example, \citet{ribeiro-etal-2020-beyond} reveal that state-of-the-art models in traditional NLP benchmarks fail simple behavioral tests of linguistic capabilities (\textit{checklists}).
\citet{chen-durrett-2019-understanding} and \citet{min-etal-2019-compositional} show that questions in multi-hop reasoning datasets such as HotpotQA by \citet{yang-etal-2018-hotpotqa} do not necessarily require multi-hop reasoning across multiple paragraphs.

To investigate how to collect high-quality, challenging questions through crowdsourcing, \citet{nangia-etal-2021-ingredients} compare different sourcing protocols and find that training workers and providing feedback about their submissions improve the difficulty and quality of their reading comprehension questions.
To encourage workers to write difficult examples, \citet{bartolo-etal-2020-beat} propose to collect questions using a model-in-the-loop setting.
Although this adversarial approach enables us to collect challenging questions efficiently, \citet{gardner-etal-2020-evaluating} point out that the collected examples might be biased towards the quirks of the adversary models.
\citet{bowman-dahl-2021-will} extend this argument, and point out that adversarial methods can systematically eliminate coverage of some phenomena.
This is also supported by \citet{kaushik-etal-2021-efficacy}, but their findings are limited to extractive QA for Wikipedia articles.
Our motivation is to see if this argument is applicable to the multiple-choice format with a wide range of passage sources for which we expect crowdworkers to write linguistically diverse questions and answer options.

\paragraph{Sources of NLU Datasets}
Reading comprehension datasets are often constructed with a limited number of passage sources.
\citet{rajpurkar-etal-2016-squad} sample about five hundred articles from the top 10,000 articles in PageRank of Wikipedia.
Similarly, \citet{dua-etal-2019-drop} curate passages from Wikipedia articles containing numeric values to collect questions for mathematical and symbolic reasoning.
\citet{khashabi-etal-2018-looking} construct a dataset in which questions are written for various passage sources such as news articles, science textbooks, and narratives.
However, we cannot use their questions for our analysis of the variation of naturally written questions because they are designed to require local multi-sentence reasoning (such as coreference resolution and paraphrasing) by filtering out questions answerable only with a single sentence.

Similarly to our work, \citet{sugawara-etal-2017-evaluation} find that readability metrics and question difficulty do not correlate in reading comprehension datasets.
Our study differs in the following two points, which could cause different findings:
First, their observational study of existing datasets has fundamental confounding factors because the questions they examine are constructed using different sourcing methods (e.g., automatic generation, expert writing, and crowdsourcing), which could have an impact on the question difficulty.
We aim to investigate uniformly crowdsourced examples across seven different sources to obtain insights for future data construction research using crowdsourcing.
Second, they define question difficulty using human annotations alone, but this does not necessarily reflect the difficulty for current state-of-the-art models.
In this study, we define the question difficulty as the human--machine performance gap using eight recent strong models, which enables a more fine-grained analysis of the collected questions for a better benchmark of current models.

\citet{fisch-etal-2019-mrqa} propose a shared task consisting of different in-domain and out-domain datasets.
However, they combine datasets in different task formats and sourcing methods, which prevents us from comparing questions across passage sources alone.
In contrast, our focus is to compare questions collected by crowdsourcing for the same task format to analyze the question difficulty for current state-of-the-art models.
We adopt the multiple-choice format because, as discussed by \citet{huang-etal-2019-cosmos}, it allows us to evaluate both human and machine performance easily.

\section{Crowdsourcing Tasks}

This study aims to analyze what kinds of passages make crowdsourced reading comprehension questions difficult.
We use Amazon Mechanical Turk. 
To collect difficult and high-quality examples, we require crowdworkers to take a qualification test \mbox{before} accepting our question writing and validation tasks.

\subsection{Worker Qualification}
\label{sec:qualification}

The qualification test has two parts, which we run in separate tasks: question answering and writing.
To take the qualification test, workers have to meet the following minimum qualifications: based in the United States, Canada, or United Kingdom, have an approval rate of at least 98\%, and have at least 1,000 approved tasks.

The question answering task is used to identify workers who answer reading comprehension questions carefully.
A single question answering task has five questions that are randomly sampled from the validation set of ReClor in which most questions are taken from actual exams.
Those who correctly answer at least four out of the five questions proceed to the next qualification phase.

The question writing task is used to familiarize workers with the writing of multiple-choice reading comprehension questions and select those who can carefully write examples.
We ask workers to write two questions given two different passages randomly sampled from the validation set of RACE \cite{lai-etal-2017-race}.
This dataset consists of self-contained passages written for middle- and high-school exams in various subjects, which we expect the workers to be able to write questions for easily.
Following \citet{nangia-etal-2021-ingredients}, we then review the workers' submissions and grade them using a rubric with four criteria: the question (1) is answerable without ambiguity (\textit{yes} or \textit{no}); (2) requires reading the whole passage (five-point scale); (3) is creative and non-obvious (five-point scale); and (4) has distractor answers that could look correct to someone who has not read the passage carefully (\textit{more than one}, \textit{one}, or \textit{no}).
We rank workers using this rubric and allow approximately the top 50\% of workers to proceed to the main writing task.
We make sure that these workers write two unambiguous and answerable questions.

\subsection{Writing Task}
\label{sec:writing}

In the main writing task, a worker is shown a single passage and asked to write a question about it along with four answer options.
We provide instructions where we describe that questions have to be challenging but still answerable and unambiguous for humans, and we include good and bad examples to illustrate what kinds of questions we aim to collect.
For example, good examples require reading the whole passage and ask about characters' motivations or consequences of described events, while bad examples only ask about a simple fact or are answerable without reading the passage (Appendix~\ref{app:instructions}).

Each worker who passes the qualification round is randomly assigned to either standard or adversarial data collection.
In the standard collection, we accept workers' submissions without any filtering.
In the adversarial collection, a written question is sent to a reading comprehension model immediately.
If the model cannot answer that question correctly, we accept it.
We allow workers to submit questions (i.e., get paid) after three attempts even if they keep failing to fool the model.
We use UnifiedQA 3B v2 \cite{khashabi-etal-2020-unifiedqa} for the adversary model, which is trained on a wide variety of question answering datasets such as MCTest, RACE, NarrativeQA \cite{kocisky-etal-2018-narrativeqa}, and SQuAD.
While the source of training data that we use in our models will inevitably influence our findings, focusing on a model with very diverse pretraining and fine-tuning will minimize this effect.

\paragraph{Passage Sources}

We use passages from the following seven sources: (1) MCTest children's narratives, (2) Project Gutenberg narratives, (3) Slate online magazine articles from the 1990s sourced from the Open American National Corpus \cite{ide-suderman-2006-integrating}, (4) middle- and high-school exams from RACE, (5) graduate-level exams from ReClor, and (6) science and (7) arts articles from Wikipedia.
We use the passages from the training sets of MCTest, RACE, and ReClor.
For Gutenberg, Slate, and Wikipedia, we split available books and articles into passages. 
Details are in Appendix~\ref{app:passage-source}.
In the writing task, a passage is randomly taken from a passage pool in which there are the same number of passages extracted from each source.

\subsection{Validation Task}
\label{sec:validation}

We collect the votes of five workers for each of the collected questions.
Those workers who passed the question answering task of the qualification round can accept the validation tasks.
To incentivize workers, we use preexisting gold-labeled examples  \citep[from][]{nangia-etal-2021-ingredients} as catch trials, representing about 10\% of the tasks, and pay a bonus of \$0.50 USD if a worker can answer those questions correctly at least 80\% of the time.
If a worker fails to answer them at least 60\% of the time, we disqualify the worker from future rounds of data collection.

\paragraph{Worker Pay and Logistics}

For the writing tasks, the base pay is \$2.00 per question, which we estimate to be approximately \$15.00 per hour based on measurements from our pilot runs.
If a worker succeeds in fooling the model in adversarial data collection, they receive an additional bonus of \$1.00.
For validation, a single task consisting of five questions pays \$2.00, which we estimate to be approximately \$15.00 per hour as well.

\section{Crowdsourcing Results}

\subsection{Dataset Construction}

We collect a total of 4,340 questions, with 620 in each of the seven sources, further divided into 310 each for the standard and adversarial methods. Each passage is paired with only one question.
We randomly sample two out of five validation votes to validate the collected examples and use the remaining three votes for measuring human performance.
In the validation, we regard a question as valid if at least one of the two votes is the same as the writer's gold answer.
If both votes are the same as the gold answer, the question is regarded as a high-agreement example.
We find that 90.3\% of the collected questions are valid (92.0\% for standard collection and 88.7\% for adversarial collection).
In addition, 65.7\% of the collected questions are classified as high-agreement (68.7\% and 62.7\% for standard and adversarial collection, respectively).
We present the dataset and worker statistics in Appendices~\ref{app:dataset} and \ref{app:worker}.

\begin{table*}[t]
\centering
\small
\def\arraystretch{1.1}
    \begin{tabular}{l@{\hspace{0.8\tabcolsep}}l@{\hspace{0.7\tabcolsep}}rrrrr@{\hspace{1.0\tabcolsep}}rrrrr}
    \toprule
    \input{main_result}
    \bottomrule
    \end{tabular}
    \caption{
        Accuracy of humans and models and the difference ($\Delta$) between human accuracy and the average zero-shot performance of eight different models (\textit{M-avg}.) for all valid questions and the high-agreement portion of them.
        The highest and lowest gaps are highlighted in bold and underlined.
        The questions are crowdsourced with (\textit{Adv.}) and without (\textit{Dir.}) adversarial feedback.
        \textit{UniQA} is the zero-shot performance by the UnifiedQA 3B model used in the adversarial data collection.
        \textit{DeBERTa} is the performance by the xlarge model fine-tuned on RACE.
    }
    \label{tbl:main}
\end{table*}

\subsection{Human Performance}

Table~\ref{tbl:main} displays human and model performance.
We use the questions that are validated using two out of five human votes in the validation step above and take the majority vote of the remaining three votes to measure human performance on them.
We observe 3.3\% and 2.0\% gaps between the standard and adversarial collection in the valid and high-agreement questions, respectively.

\subsection{Machine Performance}

To establish the model performance that is not biased towards a single model, we compute the average accuracy (\textit{M-avg.}) of eight different models from the following two classes:
RoBERTa large \citep[four models with different random seeds;][]{liu2019roberta} and DeBERTa large and xlarge \citep[v2;][]{he2020deberta} either fine-tuned on MNLI \cite{williams-etal-2018-broad} first or not.

The RoBERTa and DeBERTa models are all fine-tuned on RACE.
Among these models, DeBERTa xlarge (MNLI-fine-tuned) performs best on RACE, achieving 86.8\% accuracy.
Because UnifiedQA 3B (72.3\% on RACE) is used in the adversarial data collection, it shows lower accuracy on the adversarial questions (not included in the average).
The performance of these two models is shown for comparison in Table~\ref{tbl:main}.
Except where noted, we do not train the models on any collected questions. 

\paragraph{Supervised Performance}
For each dataset, we evaluate the performance of DeBERTa large trained on the datasets other than the target dataset in a leave-one-out manner.
Our motivation is to see whether the accuracy values significantly improve by training (i.e., the human--model gaps decrease).
If there is a large gain, it would imply that the datasets have simple patterns among examples that the models can exploit.
The results show no significant gains in the adversarial datasets, but the standard datasets show some small gains (Appendix~\ref{app:supervised}).

\begin{figure*}[t!]
    \begin{minipage}{0.33\textwidth}
        \includegraphics[width=\linewidth]{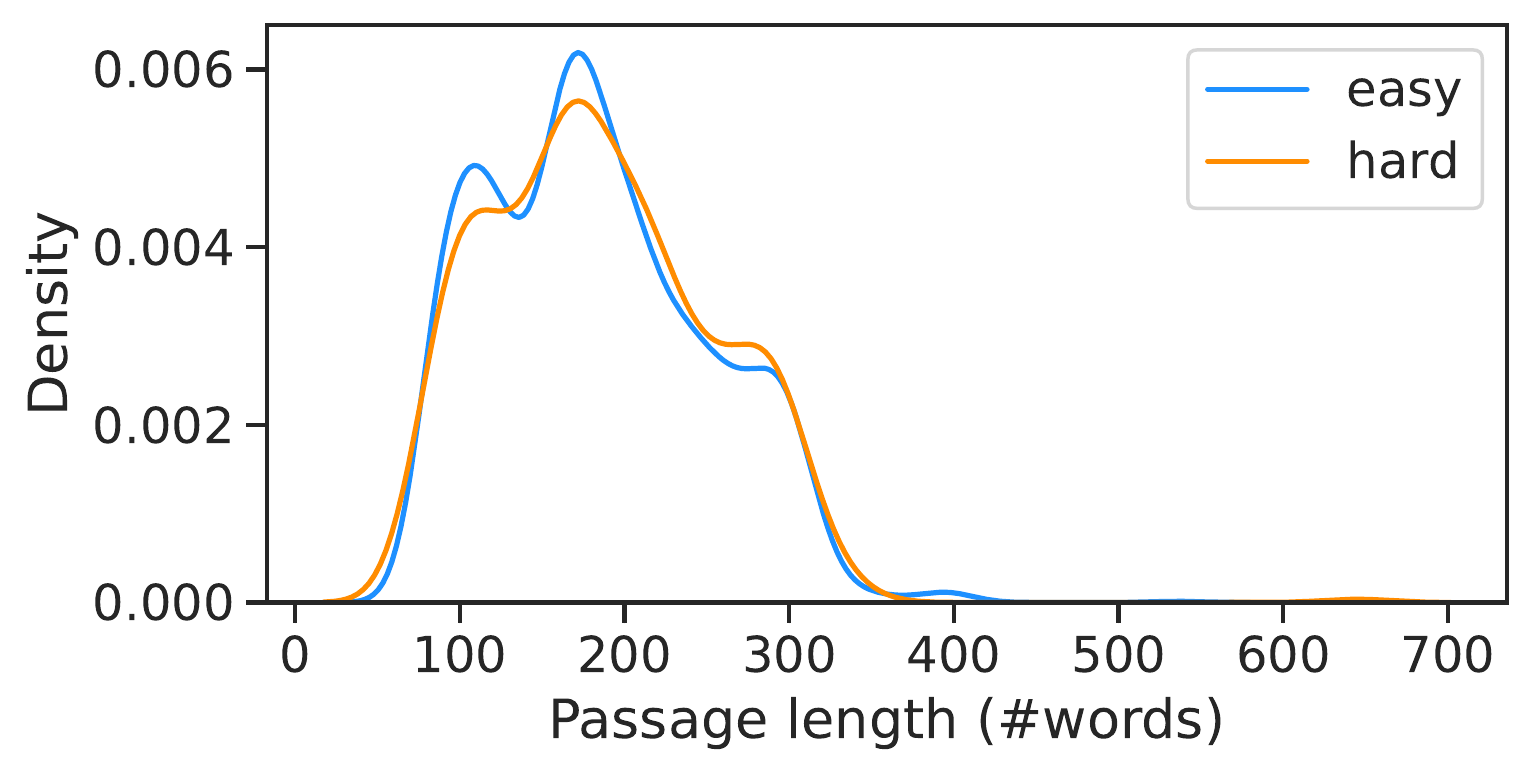}
    \end{minipage}
    \begin{minipage}{0.33\textwidth}
        \includegraphics[width=\linewidth]{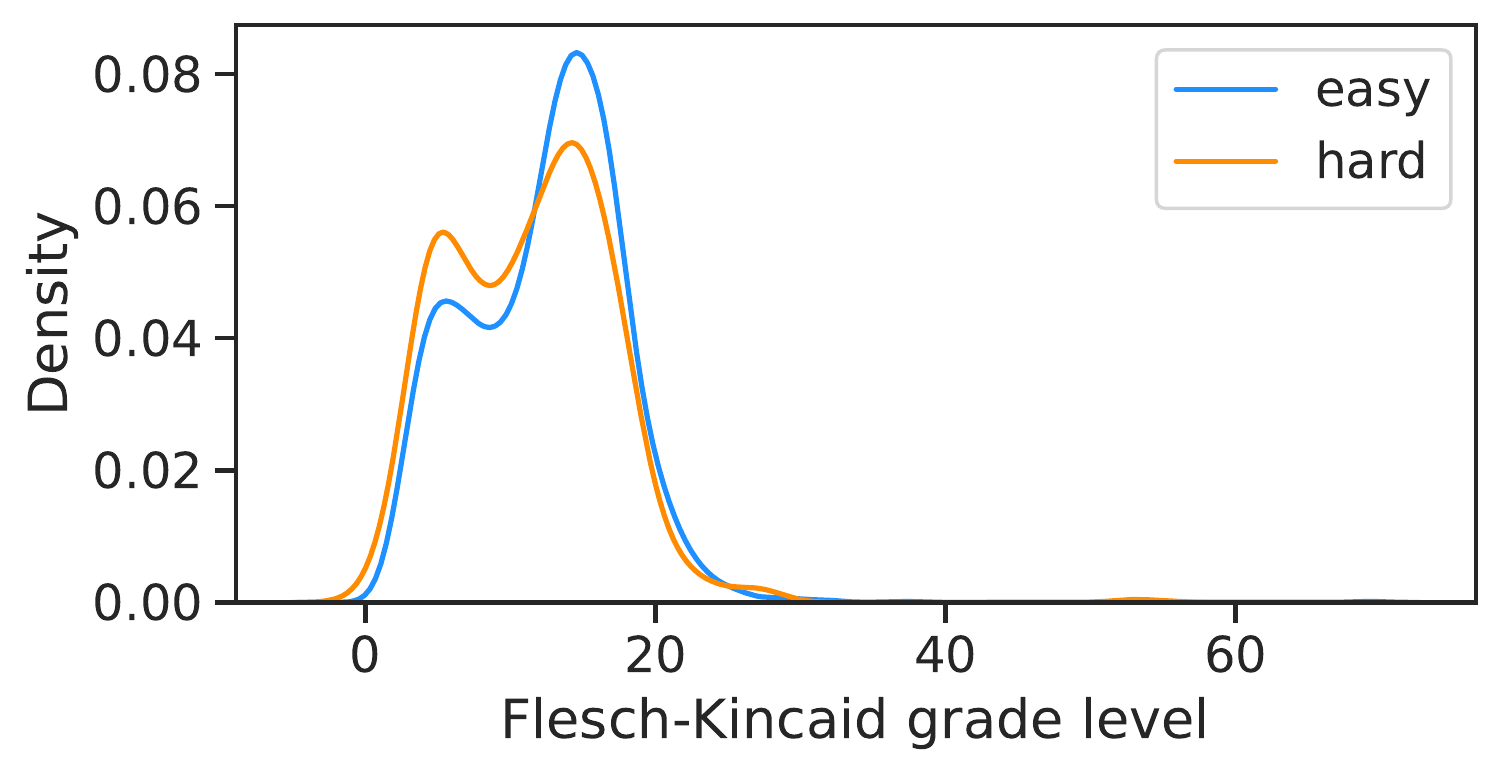}
    \end{minipage}
    \begin{minipage}{0.33\textwidth}
        \includegraphics[width=\linewidth]{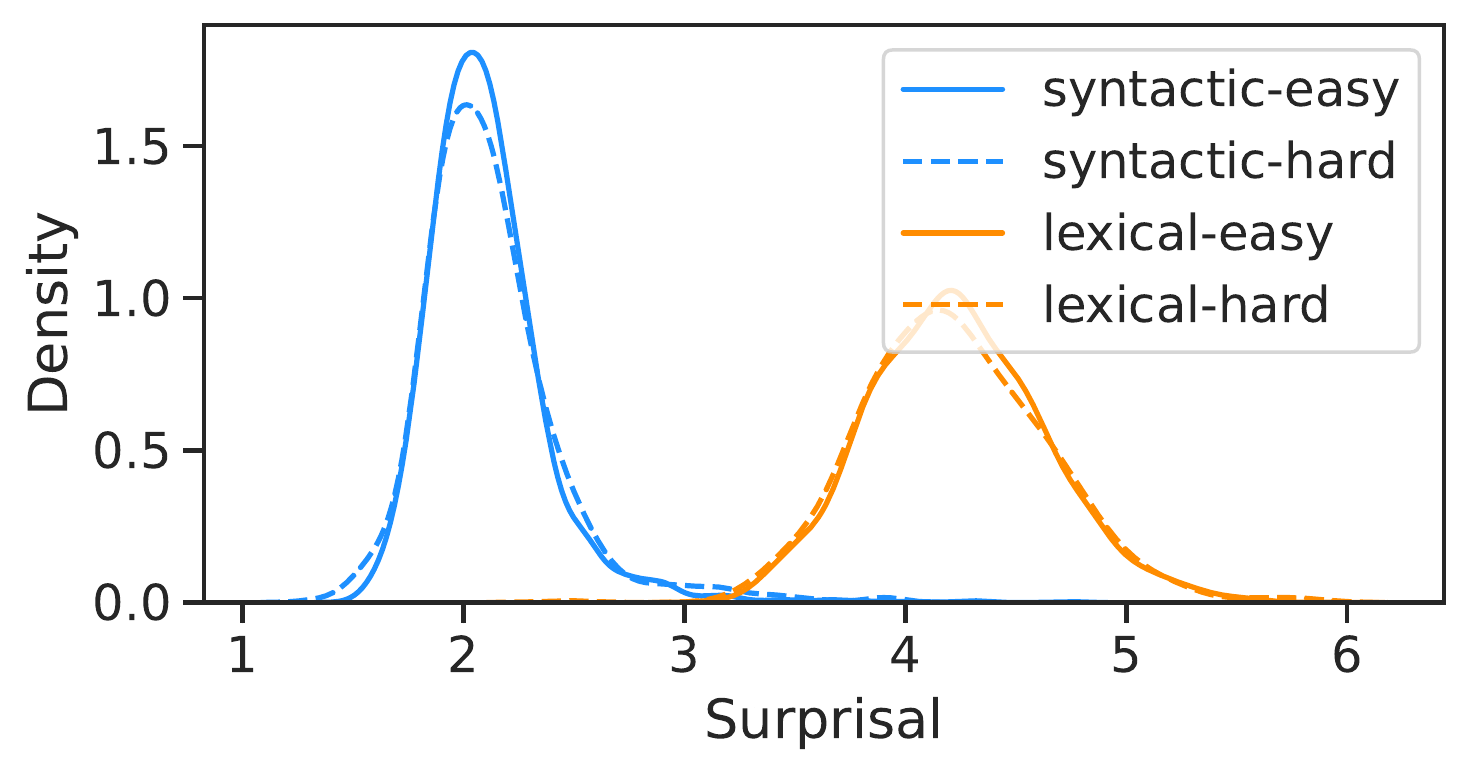}
    \end{minipage}
    \begin{minipage}{0.33\textwidth}
        \includegraphics[width=\linewidth]{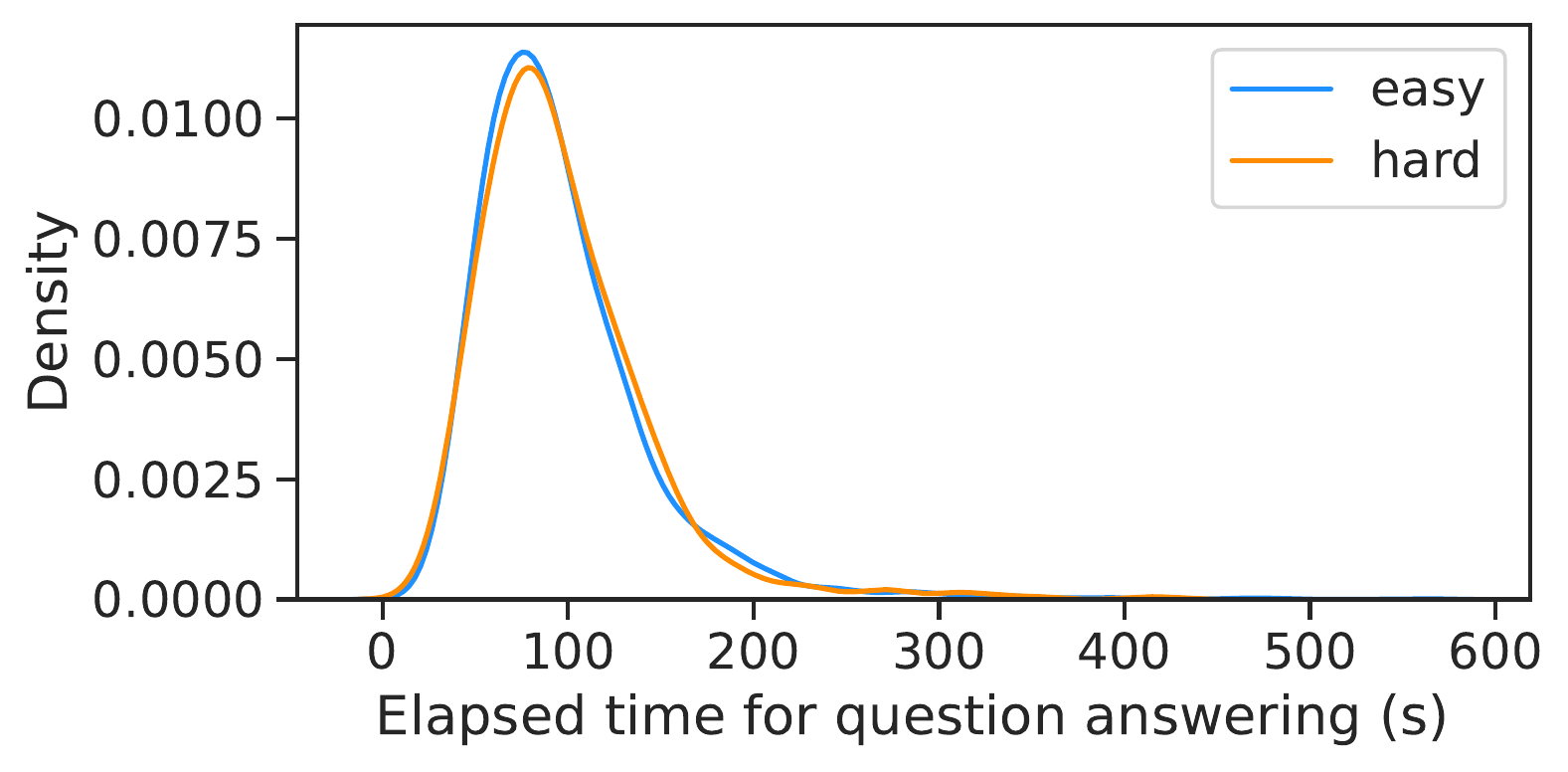}
    \end{minipage}
    \begin{minipage}{0.33\textwidth}
        \includegraphics[width=\linewidth]{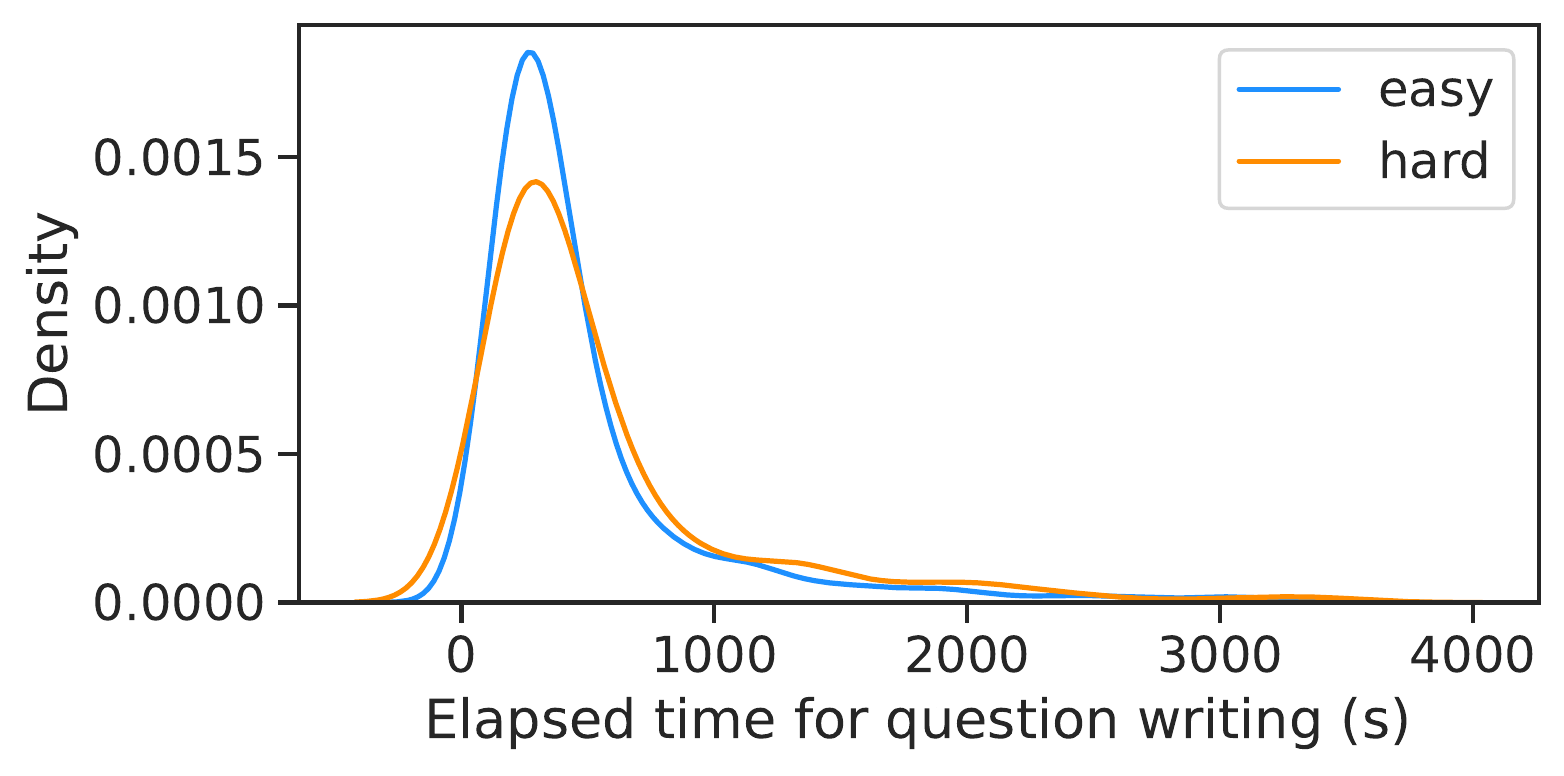}
    \end{minipage}
    \begin{minipage}{0.33\textwidth}
        \includegraphics[width=\linewidth]{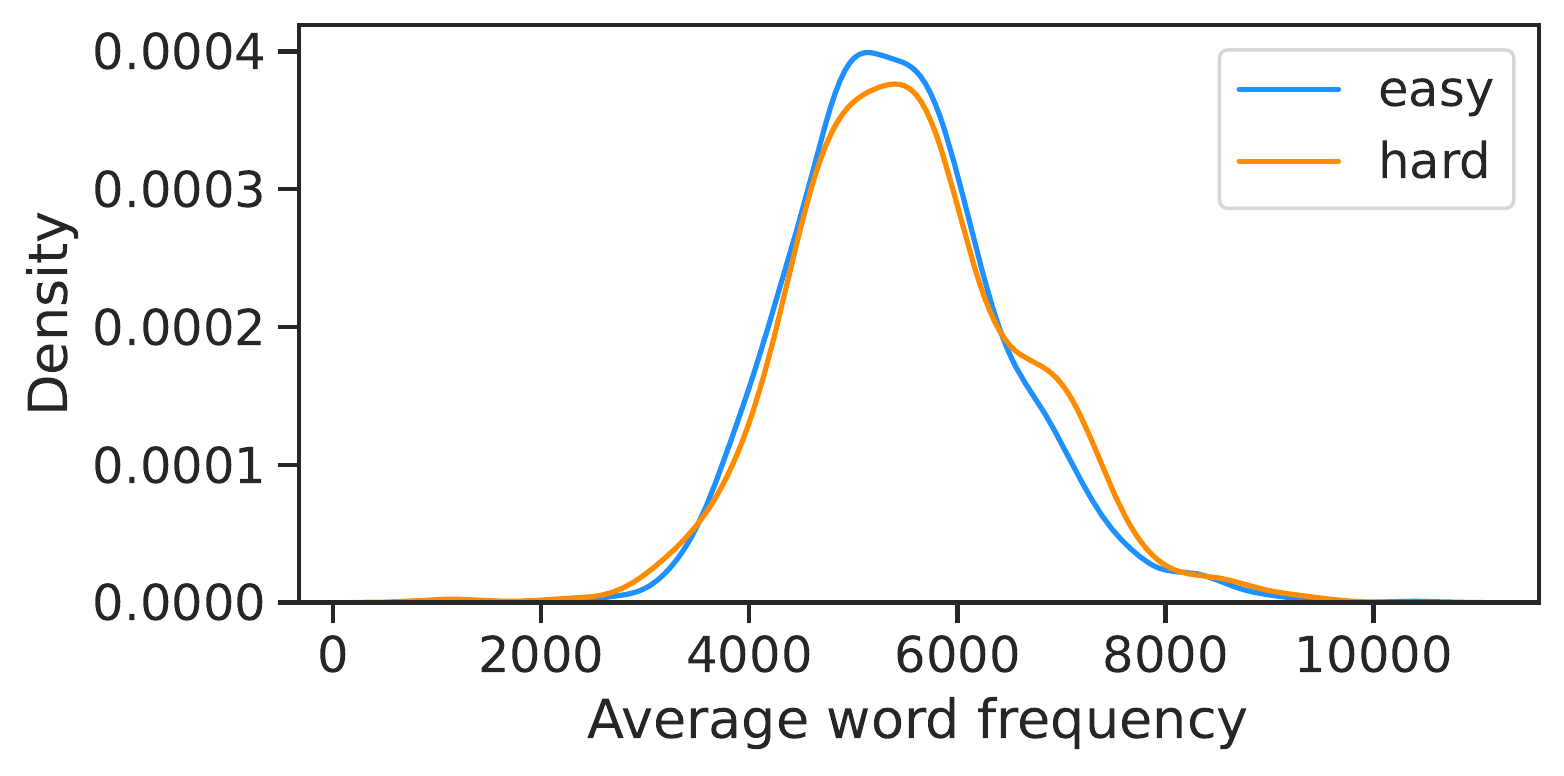}
    \end{minipage}
    \caption{Passage length, Flesch--Kincaid grade level, syntactic and lexical surprisal, elapsed time for question answering and writing, and average word frequency of passages in the easy and hard examples.}
    \label{fig:plots}
\end{figure*}

\paragraph{Partial-Input Performance}
As \citet{kaushik-lipton-2018-much} point out, reading comprehension datasets might have annotation artifacts that enable models to answer questions without passages or question sentences.
To investigate such artifacts in our collected examples, we evaluate the performance of two DeBERTa models (xlarge and large fine-tuned on MNLI), which are stronger than the others, with the ablation of questions (\textit{P+A}), passages (\textit{Q+A}), and both questions and passages (\textit{A only}).
We see large drops in the zero-shot performance of DeBERTa xlarge.
In addition, we do not observe a significant performance improvement in the supervised performance by DeBERTa large (MNLI-fine-tuned).
These results demonstrate that the collected questions and answer options do not have severe annotation artifacts for any passage source (Appendix~\ref{app:partial}).

\subsection{Human--Model Performance Gap}
Following \citet{nangia-etal-2021-ingredients}, we compute the human--model performance gap ($\Delta$) between the human and the average model accuracies to estimate the difficulty of questions for models.
We observe a small variation in the gap for different passage sources in the high-agreement questions $(\Delta=14.9\pm3.6)$.
We find the highest human performance for MCTest questions in the high-agreement portion and the lowest for Gutenberg, whereas the model's highest performance is for Slate and the lowest for MCTest.
Surprisingly, the questions sourced from MCTest, which consists of simple narrative passages, show the largest gap out of all sources for the high-agreement questions.
Although ReClor consists of passages for graduate-level exams, it produces smaller gaps than RACE, which consists of passages for middle- and high-school English exams.
Gutenberg passages are written for adults, but the examples written for those passages do not show larger gaps than those for MCTest passages.
We find a trend in the human performance: the questions of easy-to-read sources (e.g., MCTest and RACE) show higher accuracy and those of difficult-to-read sources (e.g., Gutenberg and Slate) show lower, but this trend is not observed either in the machine performance or human--machine performance gap.
These observations are inconsistent with our initial expectations in the introduction.

\section{Linguistic Analysis}
\label{sec:analysis}

We analyze how the linguistic aspects of the collected examples correlate with the human--model performance gap computed in the experiments.
To get a better estimate of human performance, we use the high-agreement examples \cite{nie-etal-2020-learn}.
For ease of comparison, we split these examples into two subsets: easy ($\Delta \leq 20\%$) and hard ($\Delta \geq 40\%$). These subsets have 1,970 and 547 examples, respectively.
Appendix~\ref{app:easy-hard-subsets} provides the frequency of easy and hard examples across the passage sources and collection methods.

\subsection{Readability Measures}

We compute the correlation between the human--model performance gap and readability measures across all valid examples (Pearson's $r$ and $p$-value) and independence between the distributions of the easy and hard subsets about the measures ($p$-value in Welch's t-test). 
Figure~\ref{fig:plots} shows the density distributions of the easy and hard subsets, 
while Appendices~\ref{app:passage-length} to \ref{app:average-word-frequency} provide the plots of all valid examples.

\paragraph{Passage Length}

We use the number of words (except for punctuation) as the passage length (top left in Figure~\ref{fig:plots}).
Across all examples, we observe $r=0.01$ ($p=0.47$) (the full plot is in Appendix~\ref{app:passage-length}).
The t-test shows $p=0.51$.
We observe no relationship between the passage length and question difficulty.
We also analyze question and option length in Appendix~\ref{app:question-length}.

\paragraph{Flesch--Kincaid Grade Level}
We use the Flesch--Kincaid grade level \cite{kincaid1975derivation} as a basic metric of text readability (top center in Figure~\ref{fig:plots}).
This metric defines readability based on an approximate US grade level with no upper bound (higher is more difficult to read). It is computed for a passage using the average number of words that appear in a sentence and the average number of syllables in a word (Appendix~\ref{app:readability}).
The correlation between the grade and human--model performance gap is $r = -0.08$ ($p<0.001$) and the t-test shows $p<0.001$.
This result demonstrates that passage readability has a small negative effect on the question difficulty, perhaps pointing to an interfering effect whereby our pre-qualified \textit{human} annotators are more likely to make mistakes on more complex passages.


\paragraph{Syntactic and Lexical Surprisal}
The Flesch--Kincaid grade level only considers sentence length and the number of syllables.
To better estimate the passage difficulty in terms of the psycholinguistic modeling of human text processing, we use syntactic and lexical surprisal measures \cite{roark-etal-2009-deriving}.
These measures are computed using incremental parsing and proved to be useful for predicting human reading time.
We observe $r=0.000$ ($p=0.99$) for syntactic surprisal and $r=-0.007$ ($p=0.66$) for lexical surprisal across all examples.
We do not observe any statistically significant difference between the easy and hard subsets (syntactic $p=0.52$ and lexical $p=0.57$ in the t-test; see top right in Figure~\ref{fig:plots}).
Appendix~\ref{app:surprisal} describes details of the calculation.

\paragraph{Annotation Speed}
Inspired by the psycholinguistic study of text complexity \cite{gibson1998linguistic,lapata-2006-automatic}, we measure the average time crowdworkers spent answering questions in the validation tasks (see bottom left in Figure \ref{fig:plots}). 
This measures the elapsed time of both reading a given passage and thinking about its question, which is used as an approximation of reading time (as a proxy of text readability).
The correlation coefficient ($r=-0.06$ with $p<0.001$) and t-test ($p=0.88$) show that there is only a small negative correlation with question difficulty.
We also measure the elapsed time for writing questions as a reference (bottom center in Figure~\ref{fig:plots} and Appendix~\ref{app:elapsed-time-writing}), observing that there is no strong correlation ($r=0.02$ with $p=0.27$).

\paragraph{Word Frequencies}
Following \citet{chen-meurers-2016-characterizing}, we analyze the effect of word frequencies on text readability.
Using word frequencies per one million words in SUBTLEXus \cite{brysbaert2009moving}, we calculate the average frequency of words appearing in a passage as a measure of passage difficulty in terms of vocabulary (a lower average frequency implies greater difficult).
We do not observe any statistically significant difference by the t-test $p=0.14$ (bottom right in Figure~\ref{fig:plots}) or Pearson's $r=0.02$ with $p=0.27$ (Appendix~\ref{app:average-word-frequency}).
We observe similar trends even when using the human performance as the difficulty measure (Appendix~\ref{app:human_difficulty}).

\subsection{Question Types}

We analyze how passage sources and collection methods affect question types in this section.

\begin{figure}[t]
    \centering
    \includegraphics[width=\linewidth]{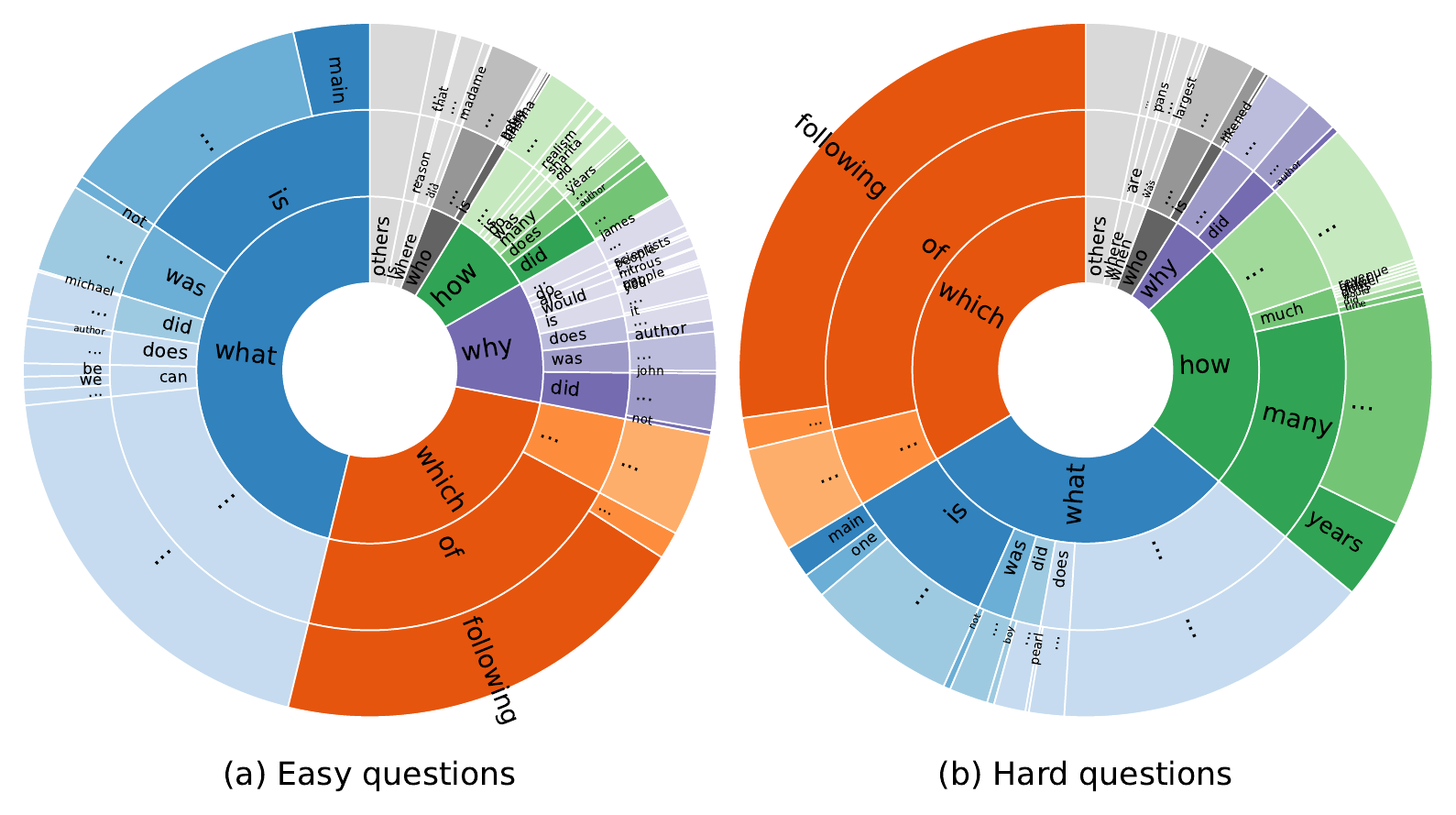}
    \caption{
        Question words and their two subsequent words in the (a) easy and (b) hard examples.
    }
    \label{fig:qtype-pie-difficulty}
\end{figure}

\paragraph{Question Words}
We automatically extract the first \textit{wh}-words that appear in each valid question;
if no \textit{wh}-word is extracted, we count the question as polar.
Figure~\ref{fig:qtype-pie-difficulty} plots the question words and their two subsequent words (except articles) in the easy and hard questions. From this we observe that the hard questions are generic, not specific to given passages (e.g., \textit{which of the following is correct?}) more often than the easy questions.
This probably results from the difference between the standard and adversarial data collection.
The workers in the adversarial collection tend to write generic questions, while those in the standard collection write questions that are more balanced (e.g., there are more easy \textit{why} and \textit{how} questions).
We also notice that the hard subset has more \textit{how many} questions.
This is likely due to the fact that it is easy for annotators to learn that numeric questions often fool the adversary model.
These observations imply that adversarial data collection tends to concentrate the distribution of questions towards a few specific question types (e.g., generic and numeric).
This is consistent with the observations in \citet{kaushik-etal-2021-efficacy}.
See Appendix~\ref{app:question-type} for details.

\begin{figure}[t]
    \includegraphics[width=\linewidth]{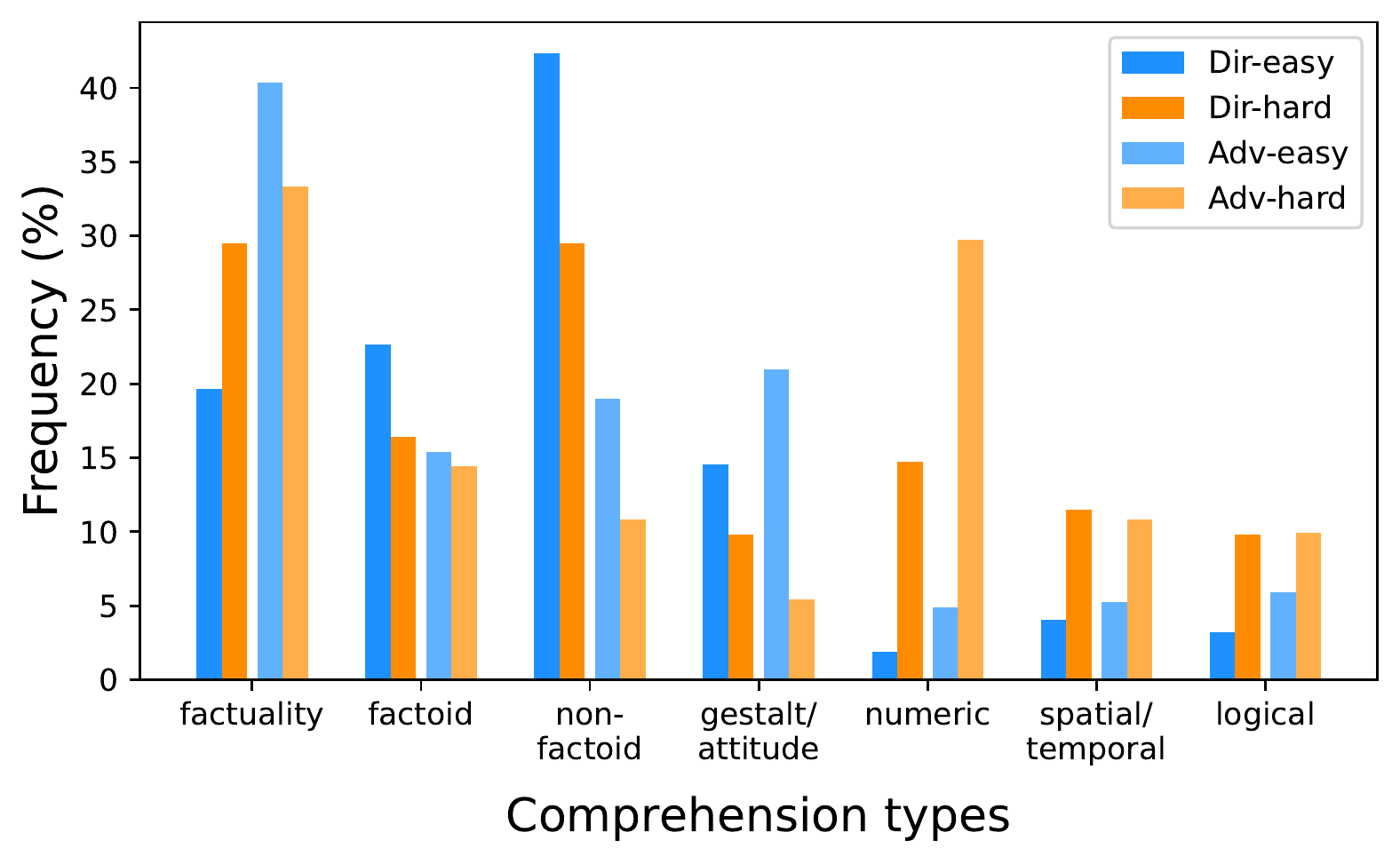}
    \caption{
        Frequency of comprehension types in the easy and hard examples for each collection method.
    }
    \label{fig:difficulty-to-reasoning}
\end{figure}

\begin{figure}[t]
    \centering
    \includegraphics[width=1.0\linewidth]{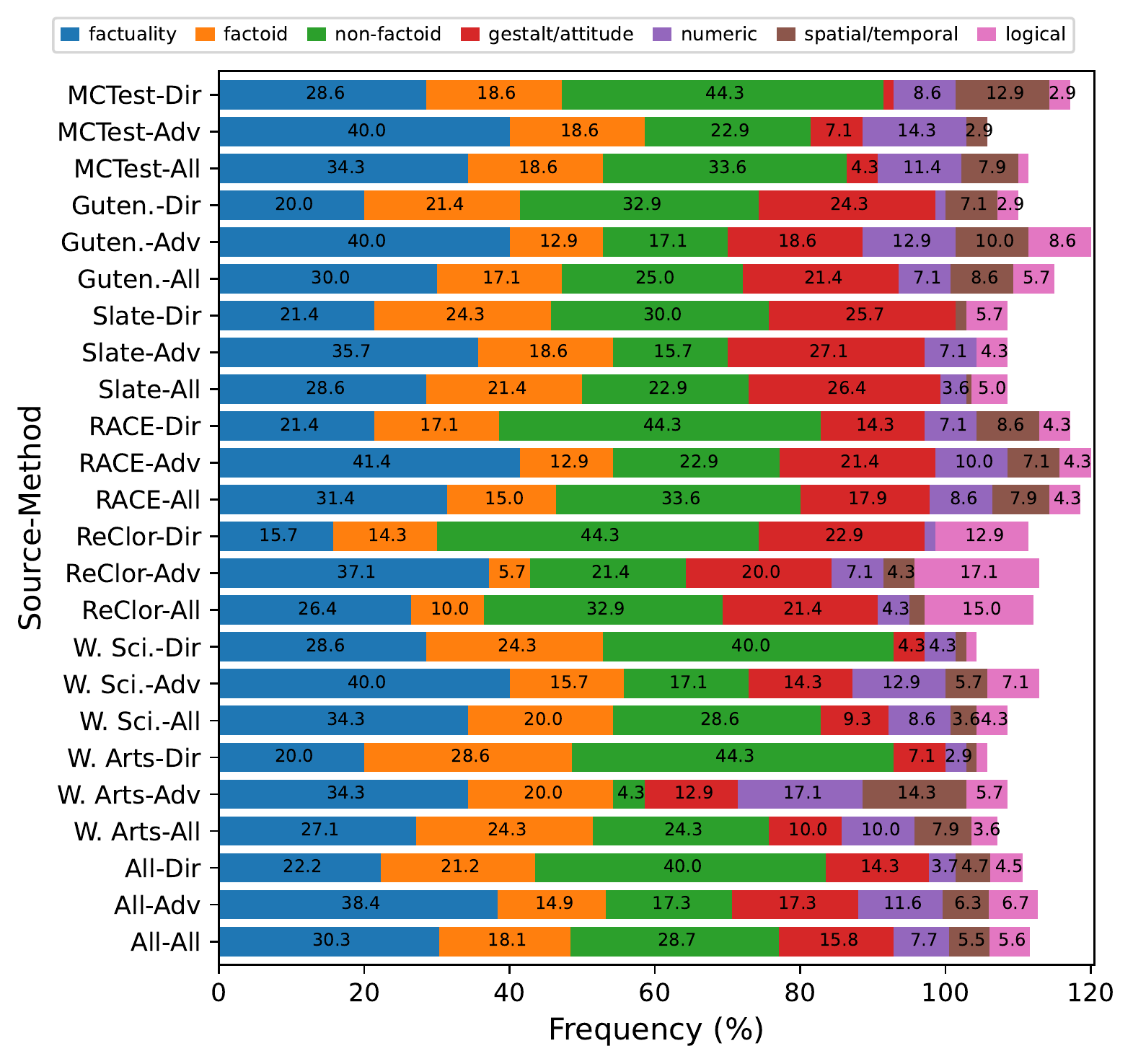}
    \caption{
        Frequency of comprehension types across passage sources and collection methods.
        Because a question can have multiple labels, the sum of the frequencies may exceed 100\%.
    }
    \label{fig:reasoning-type-detail}
\end{figure}

\paragraph{Comprehension Types}

Following \citet{bartolo-etal-2020-beat} and \citet{williams2020anlizing}, we analyze what kind of comprehension is required to answer the collected questions.
We sample a total of 980 high-agreement questions, 70 from each passage source and collection method, and then manually annotate them with one or more labels of seven comprehension types.
The definitions of these types, examples, and detailed results are presented in Appendix~\ref{app:question-type}.
Figure~\ref{fig:difficulty-to-reasoning} shows the frequency of comprehension types for different question difficulties (676 easy, 172 hard) and the collection methods.
We find that 868 questions have one label, 110 have two labels, and two have three labels.
We can see that \textit{numeric}, \textit{spatial/temporal}, and \textit{logical} questions appear more often in the hard subset in both collection methods.\footnote{In contrast, when we use the average human performance as the question difficulty measure, no comprehension type is significantly harder than the others (Appendix~\ref{app:human_difficulty}).}
Looking at the frequency across the passage sources in Figure~\ref{fig:reasoning-type-detail}, we find that there are some trends between the sources and comprehension types as follows:
\begin{itemize}
    \item Technical documents, such as those used in graduate-school-level reading comprehension exams, tend to yield logical reasoning questions (e.g., ReClor and Slate).
    \item Child-level texts tend to yield numerical reasoning questions in the standard setting (e.g., MCTest and RACE).
    In the adversarial setting, passages containing many numerical values tend to yield such questions (e.g., MCTest and Wikipedia arts).
    \item To collect gestalt questions or those considering the author’s attitude in a given passage, passages covering subjective or argumentative topics (e.g., Gutenberg, Slate, and ReClor) are suitable.
    In contrast, expository passages such as Wikipedia articles are not.
    \item Narratives and related texts (e.g., MCTest, Gutenberg, and part of RACE) involve events with characters, which tend to yield spatial/temporal reasoning questions. 
\end{itemize}
Although the definitions of our comprehension types are coarse and these trends do not ensure that specific kinds of passages always yield the target comprehension type, considering passage sources might be an effective strategy for collecting questions of an intended comprehension type.
Adversarial data collection for this purpose might not be useful because it may encourage workers to focus on writing only a few specific types of questions (e.g., numeric).

\section{Conclusion}
To make an NLU benchmark useful, it has to consist of examples that are linguistically diverse and difficult enough to discriminate among state-of-the-art models.
We crowdsource multiple-choice reading comprehension questions for passages extracted from seven different sources and analyze the effects of passage source on question difficulty and diversity.

Although we expect that the difficulty of a passage affects the difficulty of questions about that passage, the collected questions do not show any strong correlation between the human--machine performance gap and passage source, length, or readability measures.
Our manual annotation of comprehension types reveals that questions requiring numerical or logical reasoning are relatively difficult.
We also find several trends between passage sources and comprehension types.

These results suggest that when creating a new benchmark dataset, we need to select passage sources carefully, so that the resulting dataset contains questions that require an understanding of the linguistic phenomena that we are interested in.
This is especially important in the adversarial setting because it could concentrate the distribution of questions towards a few specific question types.

\section*{Ethics Statement}
We aim to accelerate scientific progress on robust general question answering, which could translate downstream to useful tools.
We are not looking at possible sources of social bias, although this issue should be highly relevant to those considering sources to use as training data for applied systems \cite{li-etal-2020-unqovering,parrish2021bbq}.
We are using Amazon Mechanical Turk despite its history of sometimes treating workers unfairly \cite{kummerfeld-2021-quantifying}, especially in recourse for unfair rejections.
We make sure that our own pay and rejection policies are comparable to in-person employment, but acknowledge that our study could encourage others to use Mechanical Turk, and that they might not be so careful.
This work passed review or is exempt from the oversight of the internal review boards of the authors' institutes.

\section*{Acknowledgments}

We thank Saranya Venkatraman and Ethan Perez for their feedback on early drafts of this paper.
For his early contributions to this project, we thank Harsh Trivedi.
SS was supported by JST PRESTO Grant No. JPMJPR20C4. 
This project has benefited from financial support to SB by Eric and Wendy Schmidt (made by recommendation of the Schmidt Futures program), Samsung Research (under the project \textit{Improving Deep Learning using Latent Structure}), and Apple.
This material is based upon work supported by the National Science Foundation under Grant Nos. 1922658 and 2046556. Any opinions, findings, and conclusions or recommendations expressed in this material are those of the author(s) and do not necessarily reflect the views of the National Science Foundation.

\bibliographystyle{acl_natbib}
\bibliography{anthology,custom}

\appendix

\section{Passage Sources}
\label{app:passage-source}

From Project Gutenberg, we use books from the adventure, fiction, humor, novel, and story genres.\footnote{\url{https://www.gutenberg.org/}}

From Wikipedia articles, we use articles listed as Level 3 vital articles.\footnote{\url{https://en.wikipedia.org/wiki/Wikipedia:Vital_articles}} For science, we include health, medicine and disease, science, technology, and mathematics categories. For the arts, we include history, arts, philosophy and religion, and society and social sciences categories.

\section{Dataset Statistics}
\label{app:dataset}

\begin{table}[tb]
\centering
\small
    \begin{tabular}{llrrrr}
    \toprule
    \input{dataset_stats}
    \bottomrule
    \end{tabular}
    \caption{
        Frequency of valid and high-agreement examples for different passage sources and collection methods.
        }
    \label{tbl:stats}
\end{table}

Table~\ref{tbl:stats} presents the frequencies of valid and high-agreement examples across the passage sources and collection methods.

\section{Worker Statistics}
\label{app:worker}

Of the 1,050 workers who joined the question-answering phase of the qualification round, 259 workers (24.7\%) passed it.
From them, 157 workers submitted the question writing task, and 72 workers (36 each for the standard and adversarial collection) qualified for the main writing task, from which 49 workers joined.
The workers were allowed to write up to 250 questions.
A total of 167 workers participated in the validation task.
No worker answered more than 730 questions.
Data collection took approximately a month including the qualification round and the validation task.

\section{Supervised Model Performance}
\label{app:supervised}

\begin{table}[t]
\centering
\small
\begin{tabular}{llrr}
    \toprule
    \input{transfer_results}
    \bottomrule
\end{tabular}
\caption{
    Supervised performance of DeBERTa large. The accuracy of each row is given by the model trained on the questions of the other rows (leave-one-out training).
    Subscript values show the difference from its zero-shot accuracy.
    }
    \label{tbl:supervised}
\end{table}

Table~\ref{tbl:supervised} shows the supervised performance of the DeBERTa large model.

\section{Partial-Input Model Performance}
\label{app:partial}

\begin{table}[t]
\centering
\small
    \begin{tabular}{l@{\hspace{0.9\tabcolsep}}lccc}
    \toprule
    \input{partial_input}
    \bottomrule
    \end{tabular}
    \caption{
        Zero-shot performance of DeBERTa xlarge trained on RACE with ablation settings.
        We ablate questions (\textit{P+A}), passages (\textit{Q+A}), or both questions and passages (\textit{A only}) from the input.
        Subscripts show the difference from the full-input accuracy.
    }
    \label{tbl:partial-input}
\end{table}

\begin{table}[t]
\centering
\def\arraystretch{1.2}
\begin{tabular}{lccc}
    \toprule
    Method & P+A & Q+A & A only \\ \midrule
    Dir.  & 71.6 $^{\pm0.8}_{+0.6}$ & 46.0 $^{\pm2.2}_{+4.7}$ & 38.6 $^{\pm1.5}_{+5.4}$ \\
    Adv.  & 51.9 $^{\pm1.3}_{+1.2}$ & 41.5 $^{\pm2.2}_{+1.5}$ & 32.7 $^{\pm0.6}_{+3.3}$ \\
    \bottomrule
    \end{tabular}
    \caption{Supervised performance (three-fold cross validation) of DeBERTa large on the partial inputs.
    Superscripts show standard deviation and subscripts show gains over the zero-shot performance.
    }
    \label{tbl:partial-input-supervised}
\end{table}

Tables~\ref{tbl:partial-input} and \ref{tbl:partial-input-supervised} report the zero-shot performance of DeBERTa xlarge and the supervised performance of DeBERTa large (MNLI).

\section{Easy and Hard Subsets}
\label{app:easy-hard-subsets}

\begin{table}[th]
    \centering\small
    \begin{tabular}{llcc} \toprule
    Source & Method & Easy & Hard \\ \midrule
    MCTest & Dir. & 8.1 & 6.4 \\
    \rowcolor{gray!15}
    \hphantom{MCTest} & Adv. & 6.5 & 13.2 \\
    \rowcolor{gray!30}
    \hphantom{MCTest} & Total & 14.7 & 19.6 \\
    Gutenberg & Dir. & 8.1 & 4.6 \\
    \rowcolor{gray!15}
    \hphantom{Gutenberg} & Adv. & 6.2 & 7.3 \\
    \rowcolor{gray!30}
    \hphantom{Gutenberg} & Total & 14.3 & 11.9 \\
    Slate & Dir. & 8.4 & 2.9 \\
    \rowcolor{gray!15}
    \hphantom{Slate} & Adv. & 5.8 & 7.7 \\
    \rowcolor{gray!30}
    \hphantom{Slate} & Total & 14.2 & 10.6 \\
    RACE & Dir. & 8.7 & 5.7 \\
    \rowcolor{gray!15}
    \hphantom{RACE} & Adv. & 6.2 & 12.1 \\
    \rowcolor{gray!30}
    \hphantom{RACE} & Total & 14.9 & 17.7 \\
    ReClor & Dir. & 8.6 & 5.5 \\
    \rowcolor{gray!15}
    \hphantom{ReClor} & Adv. & 5.5 & 8.0 \\
    \rowcolor{gray!30}
    \hphantom{ReClor} & Total & 14.2 & 13.5 \\
    Wiki. Sci. & Dir. & 8.7 & 4.4 \\
    \rowcolor{gray!15}
    \hphantom{Wiki. Sci.} & Adv. & 5.1 & 10.2 \\
    \rowcolor{gray!30}
    \hphantom{Wiki. Sci.} & Total & 13.8 & 14.6 \\
    Wiki. Arts & Dir. & 8.3 & 3.1 \\
    \rowcolor{gray!15}
    \hphantom{Wiki. Arts} & Adv. & 5.7 & 9.0 \\
    \rowcolor{gray!30}
    \hphantom{Wiki. Arts} & Total & 14.0 & 12.1 \\ \midrule
    \# Questions & & 1,970 & 547 \\
    \bottomrule
    \end{tabular}
    \caption{Distribution (\%) of easy and hard questions from each passage source and collection method.}
    \label{tbl:easy-hard-subsets}
\end{table}

Table~\ref{tbl:easy-hard-subsets} presents the frequency of easy and hard examples across passage sources and collection methods.

\section{Passage Length}
\label{app:passage-length}

\begin{figure}[t]
    \includegraphics[width=\linewidth]{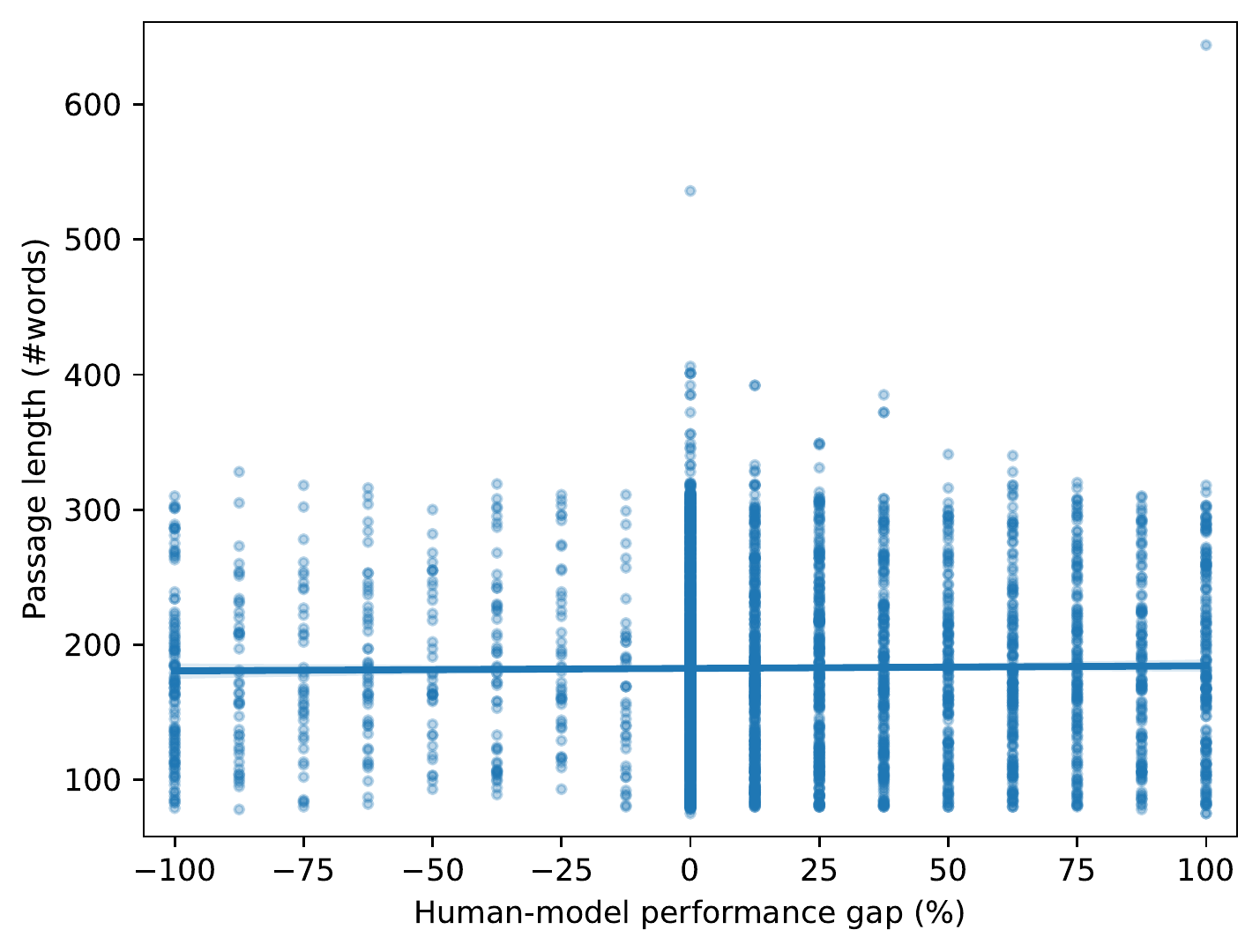}
    \caption{Passage length (number of words) and human--model performance gap. Pearson's $r=0.01$ with $p = 0.54$.}
    \label{fig:passage-length-full-plot}    
\end{figure}

Figure~\ref{fig:passage-length-full-plot} shows the relationship between the passage length and the human--model performance gap.

\section{Question and Option Length}
\label{app:question-length}

\begin{figure}[t]
    \includegraphics[width=\linewidth]{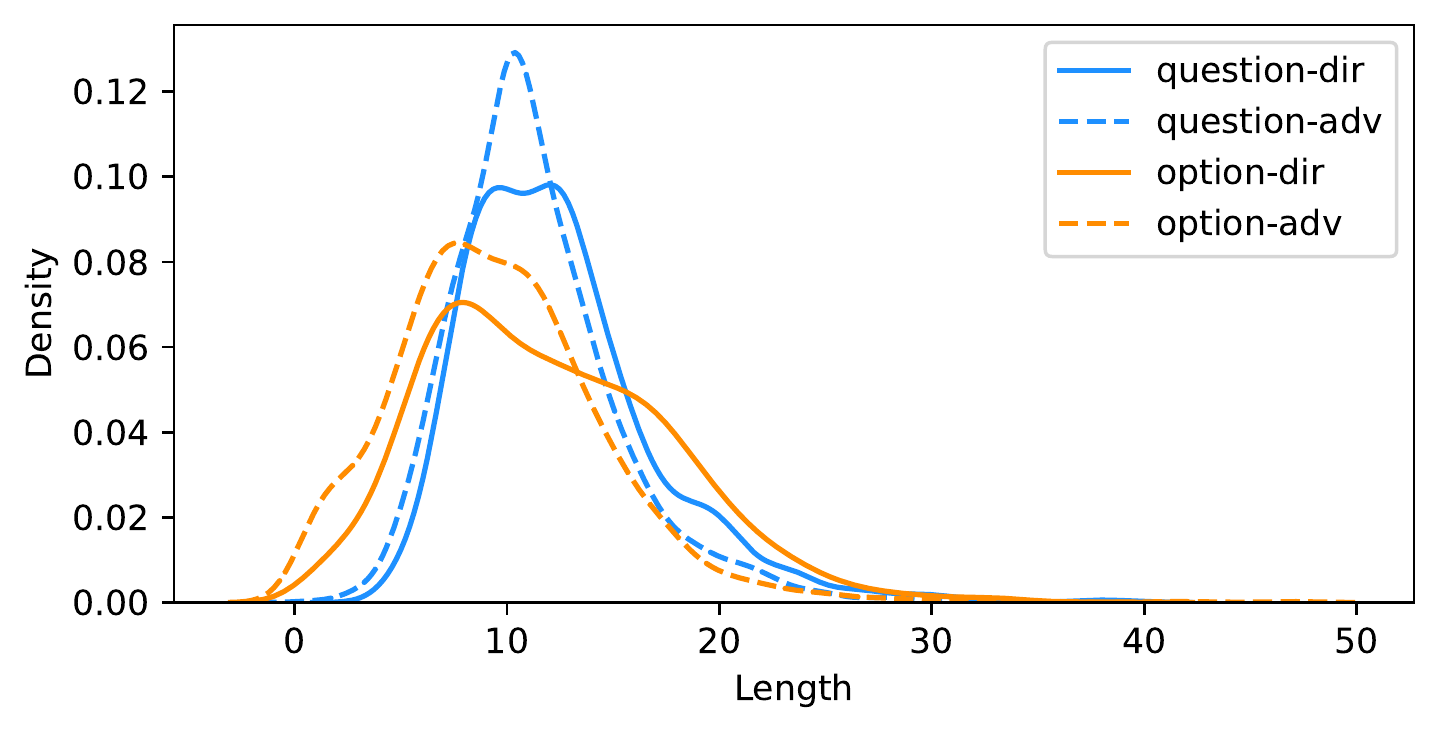}
    \caption{
        Question and option lengths (number of words) of examples collected in the standard and adversarial methods.
    }
    \label{fig:question-answer-length-plot}
\end{figure}

\begin{figure}[t]
    \includegraphics[width=\linewidth]{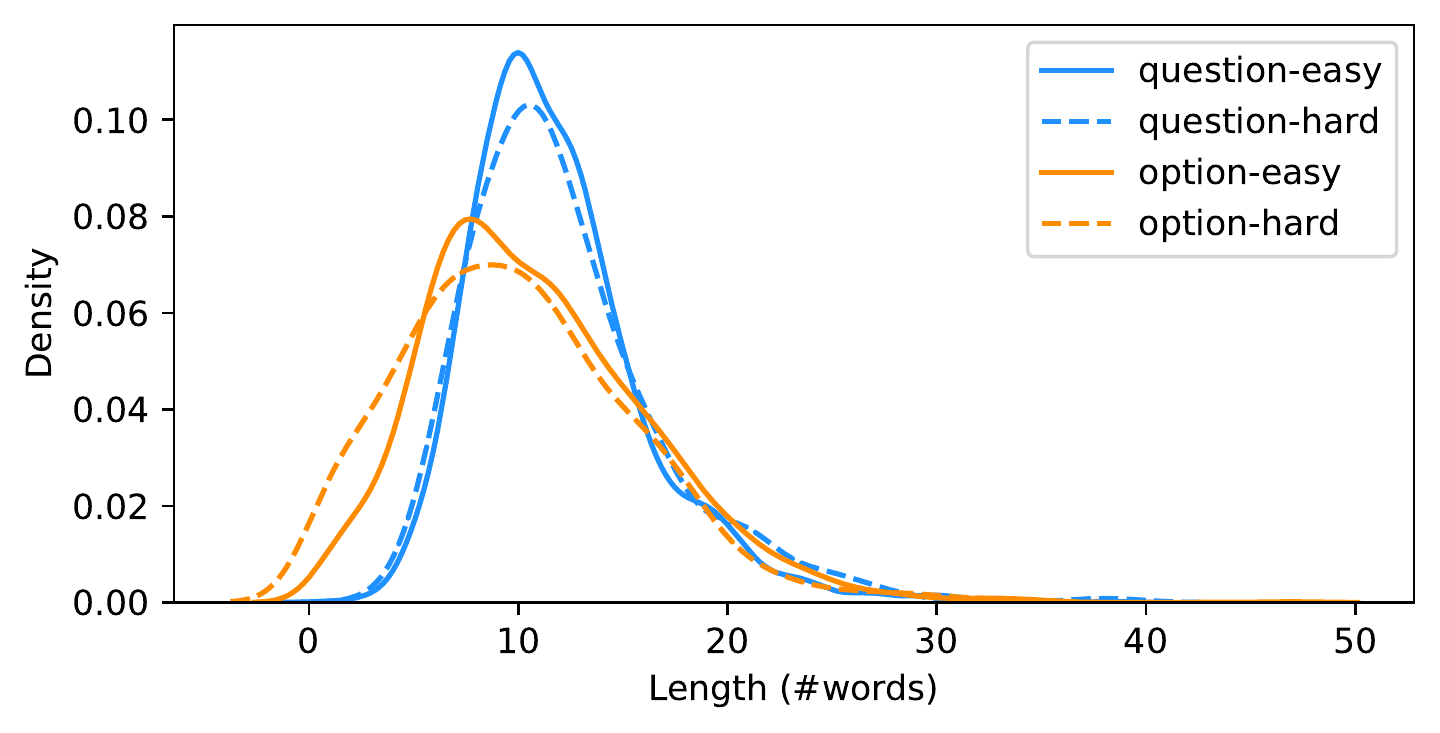}
    \caption{
        Question and option lengths (number of words) of easy and hard examples.
    }
    \label{fig:question-answer-length-gap-plot}
\end{figure}

We plot the average question and option length (the number of words except for punctuation) in the high-agreement examples in Figure~\ref{fig:question-answer-length-plot} across the collection methods and in Figure~\ref{fig:question-answer-length-gap-plot} across the easy and hard subsets.
The distributions of question and option length have slightly higher variances in the standard data collection than in the adversarial data collection.
This result is consistent with the observation in \citet{nangia-etal-2021-ingredients}.

\section{Readability Level}
\label{app:readability}

\begin{figure}[t]
    \includegraphics[width=\linewidth]{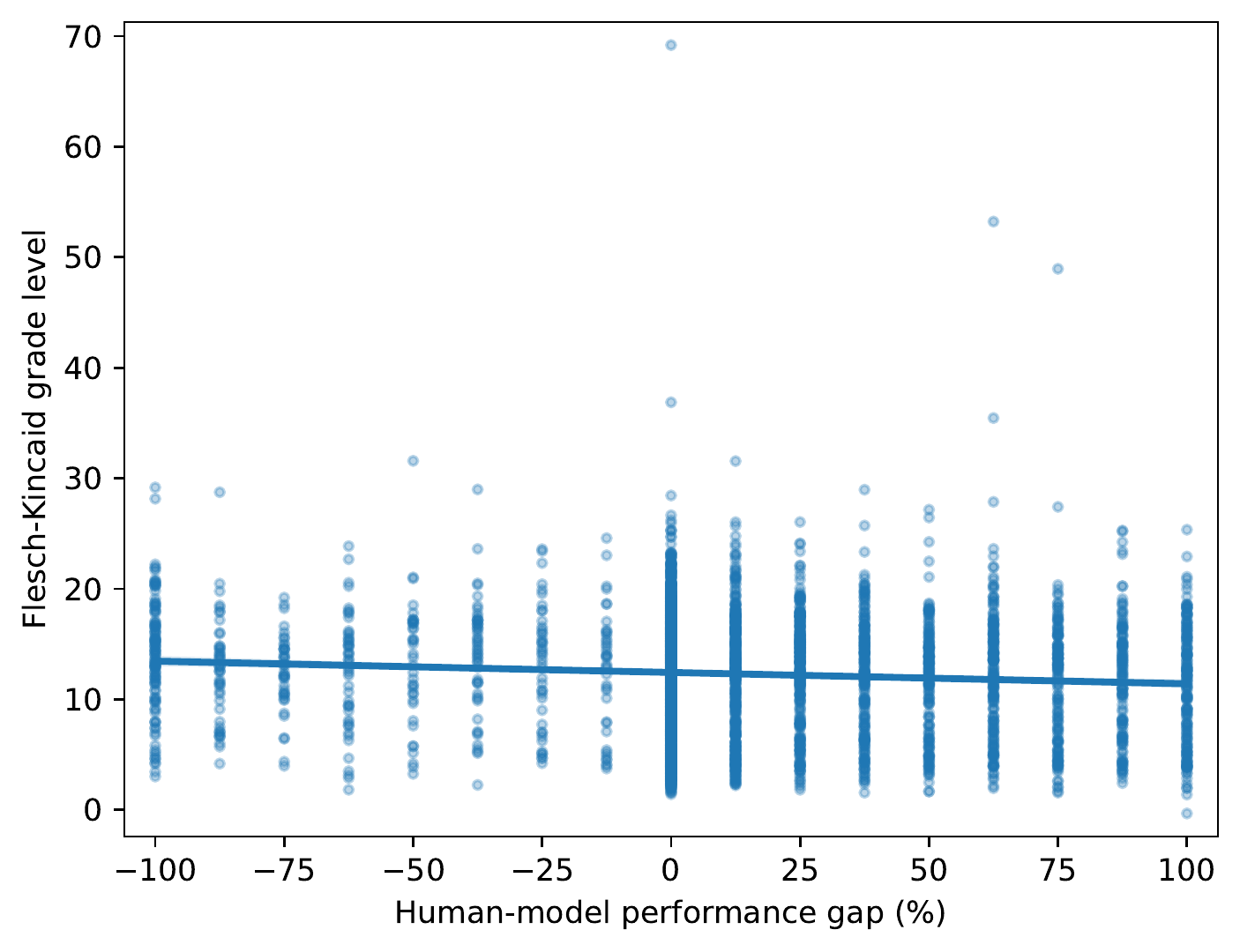}
    \caption{Flesch--Kincaid grade level and human--model performance gap. Pearson's $r=-0.08$ with $p < 0.001$.}
    \label{fig:readability-full-plot}
\end{figure}

Figure~\ref{fig:readability-full-plot} shows the plot between Flesch--Kincaid grade level \cite{kincaid1975derivation} and the human--model performance gap.
We compute the grade level ($L$) of a passage using the following formula:
\begin{equation}
    L = 0.39 * m + 11.8 * n - 15.59
\end{equation}
where $m$ is the average length of the sentences and $n$ is the average number of syllables of the words in the passage.
To estimate the number of syllables in a word, we use the implementation of the sonority sequencing principle \cite{bartlett-etal-2009-syllabification} in NLTK \cite{bird2009natural}.\footnote{\url{https://www.nltk.org/_modules/nltk/tokenize/sonority_sequencing.html}}

\section{Syntactic and Lexical Surprisal}
\label{app:surprisal}

\begin{figure}[t]
    \includegraphics[width=\linewidth]{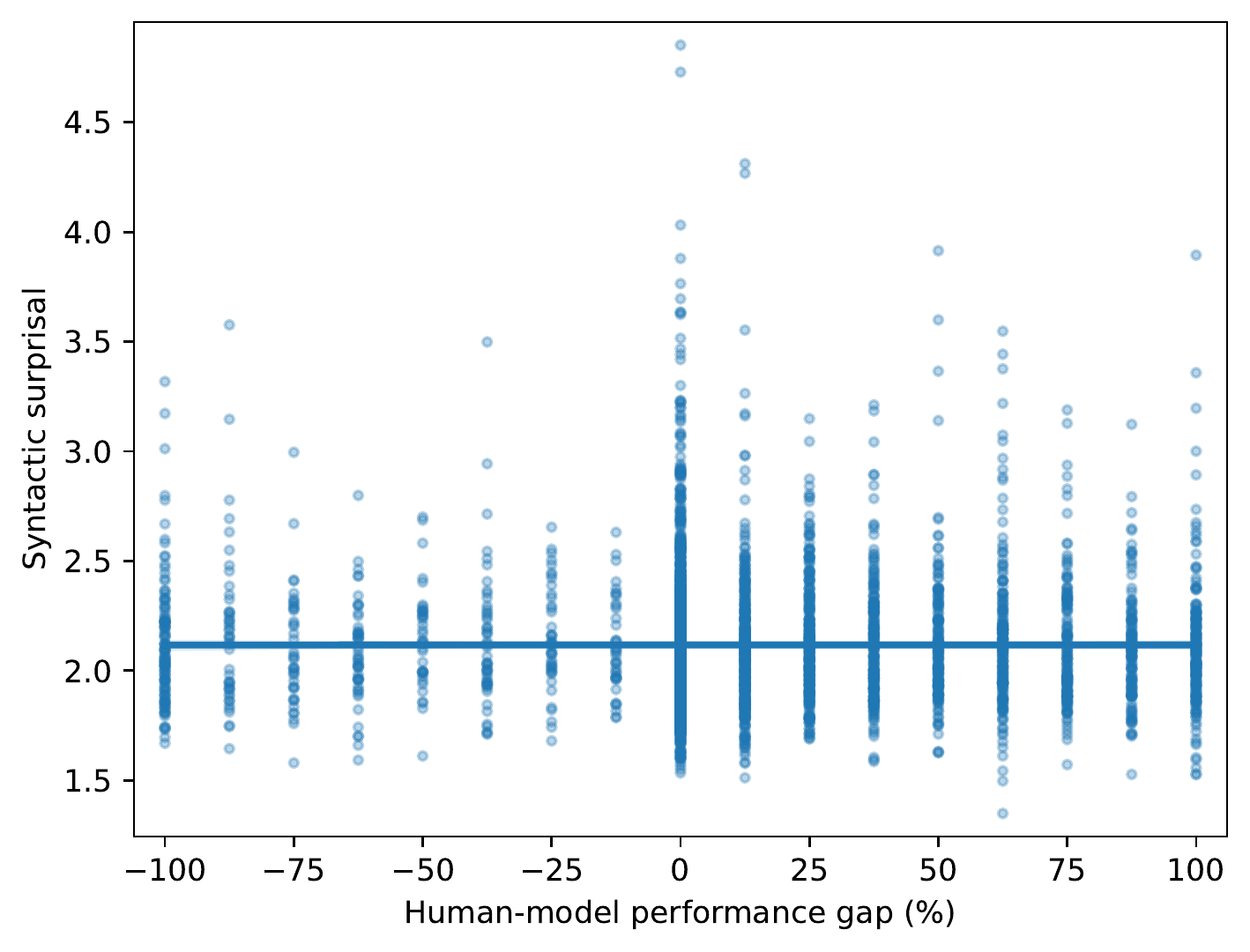}
    \caption{
        Syntactic surprisal for all valid examples. Pearson's $r=-0.003$ with $p=0.86$.
    }
    \label{fig:syntactic-surprisal-full}
\end{figure}

\begin{figure}[t]
    \includegraphics[width=\linewidth]{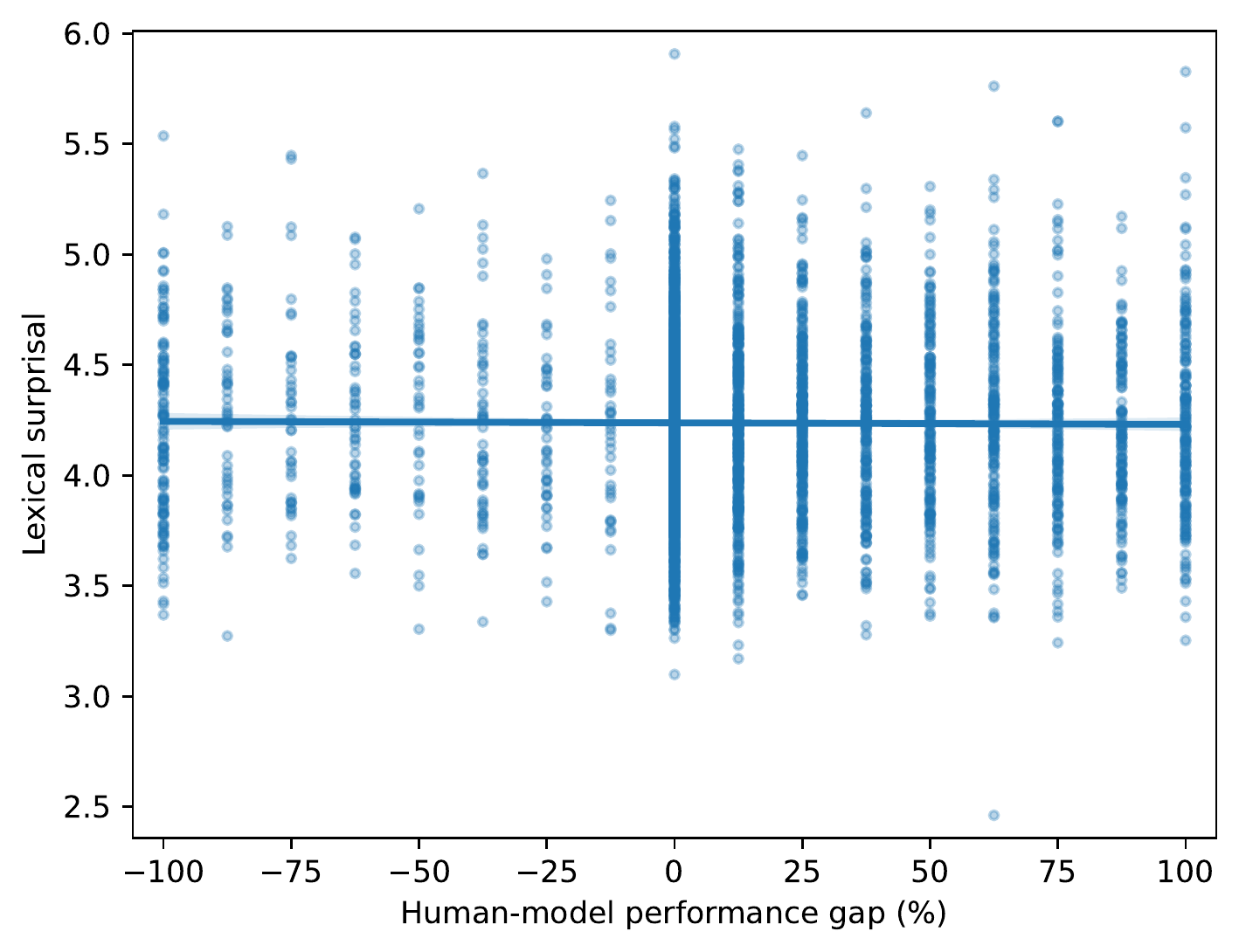}
    \caption{
        Lexical surprisal for all valid examples. Pearson's $r=-0.002$ with $p=0.90$.
    }
    \label{fig:lexical-surprisal-full}
\end{figure}

Figures~\ref{fig:syntactic-surprisal-full} and \ref{fig:lexical-surprisal-full} show syntactic and lexical surprisal measures, respectively, for all examples.
Following \citet{roark-etal-2009-deriving}, we compute a surprisal value for each word, then take the average for each sentence, and finally take the average over the whole passage.
We use an incremental parser with a lexicalized probabilistic context-free grammar.\footnote{\url{https://github.com/roarkbr/incremental-top-down-parser}}

\section{Elapsed Time for Answering Questions}
\label{app:elapsed-time-writing}

\begin{figure}[th]
    \includegraphics[width=\linewidth]{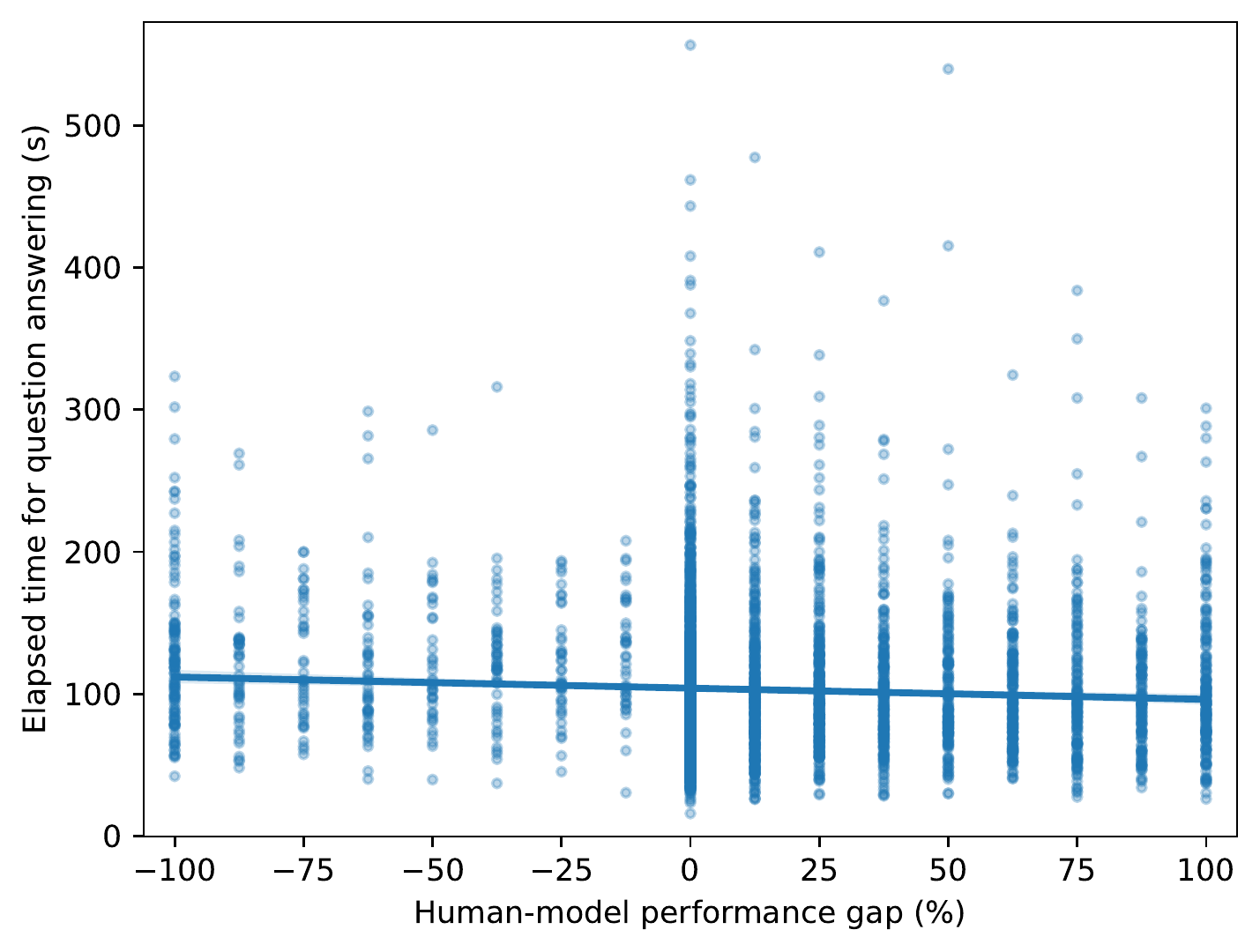}
    \caption{
        Elapsed time (s) for answering all examples. Pearson's $r=-0.08$ with $p<0.001$.
    }
    \label{fig:elapsed-time-answering-full}
\end{figure}

\begin{figure}[th]
    \includegraphics[width=\linewidth]{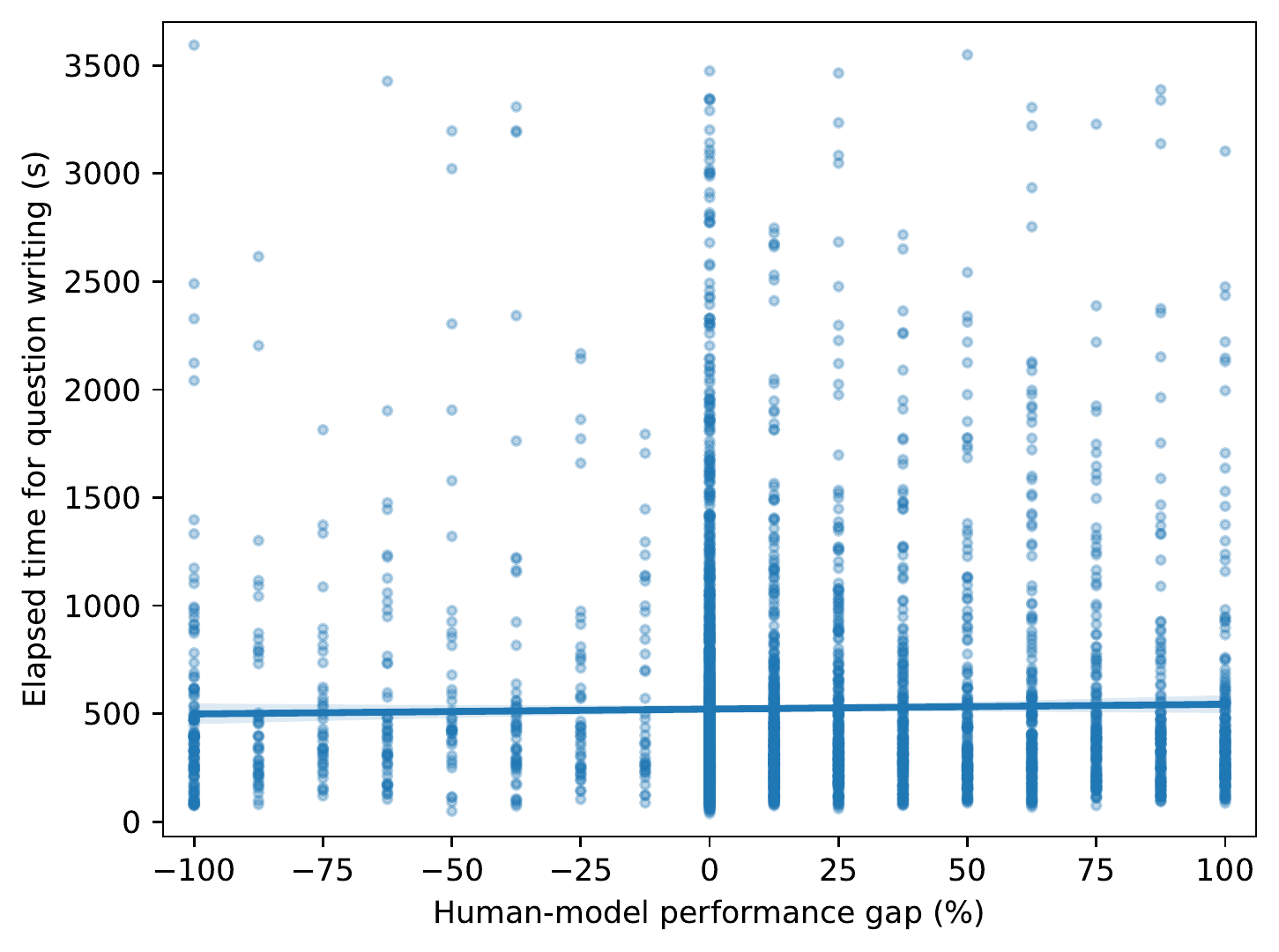}
    \caption{
        Elapsed time (s) for writing all examples. Pearson's $r=0.03$ with $p=0.03$.
    }
    \label{fig:elapsed-time-writing-full}
\end{figure}

Figure~\ref{fig:elapsed-time-answering-full} shows the plot of time elapsed by humans while answering questions in the validation task.
We measure the elapsed time from when a worker opens a task to when they submit their answer. 
In addition, we measure the elapsed time for writing questions as a reference (Figure~\ref{fig:elapsed-time-writing-full}).
We observe that workers take slightly longer to write hard examples than easy examples.

\section{Average Word Frequencies}
\label{app:average-word-frequency}

\begin{figure}[t]
    \includegraphics[width=\linewidth]{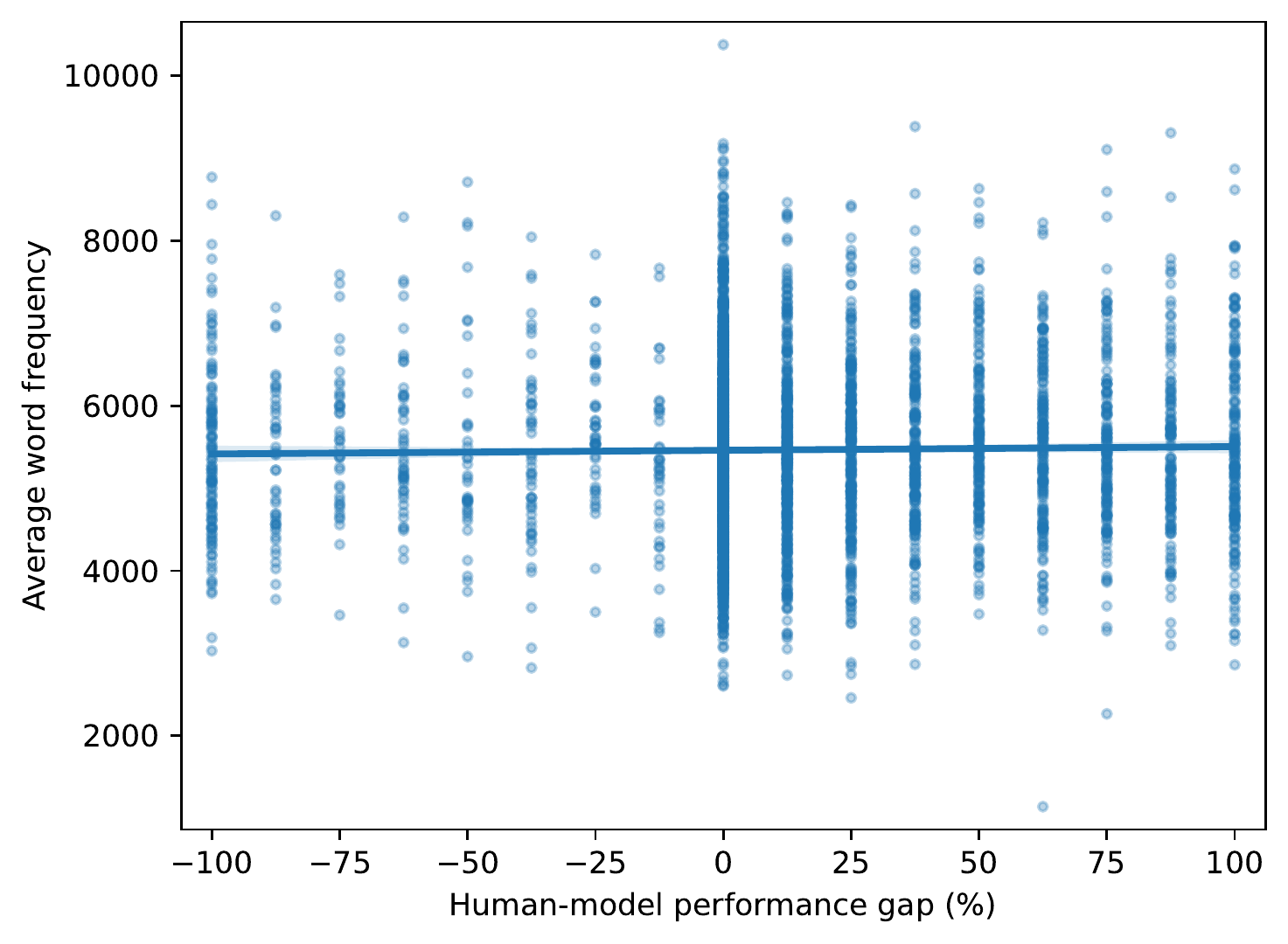}
    \caption{
        Average word frequencies using SUBTLEXus values. Pearson's $r=0.02$ with $p=0.23$.
    }
    \label{fig:average-word-frequency-full}
\end{figure}

Figure~\ref{fig:average-word-frequency-full} plots the average word frequencies of all examples.
We refer to SUBTLEXus \cite{brysbaert2009moving} for the word frequencies per one million words in a corpus of American English subtitles.

\section{Question and Comprehension Types}
\label{app:question-type}

\begin{figure}[th]
    \includegraphics[width=\linewidth]{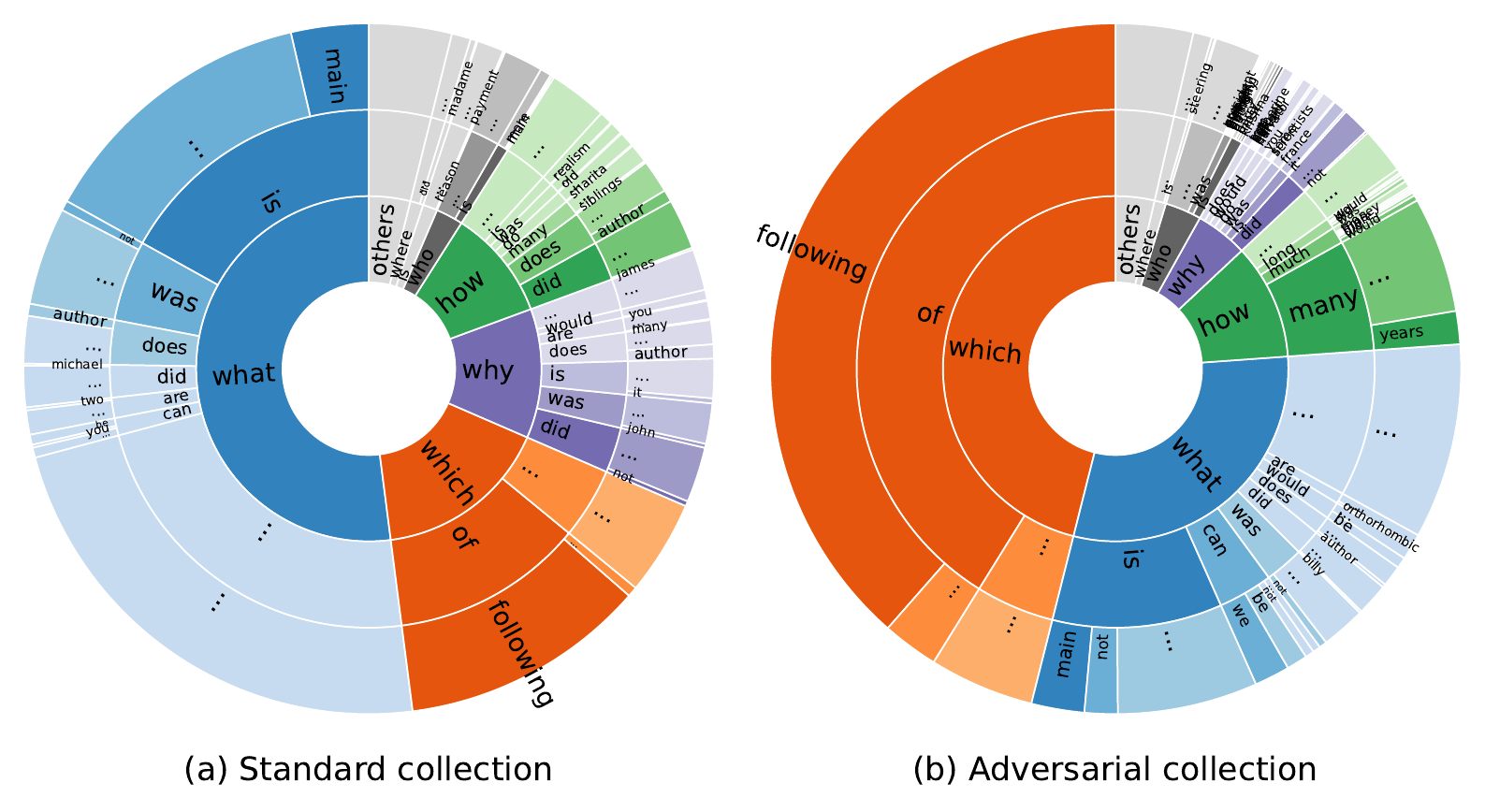}
    \caption{
        Question words and their two subsequent words in the (a) standard and (b) adversarial collection methods.
    }
    \label{fig:qtype-pie-methods}
\end{figure}

\begin{figure}[tbh]
    \includegraphics[width=\linewidth]{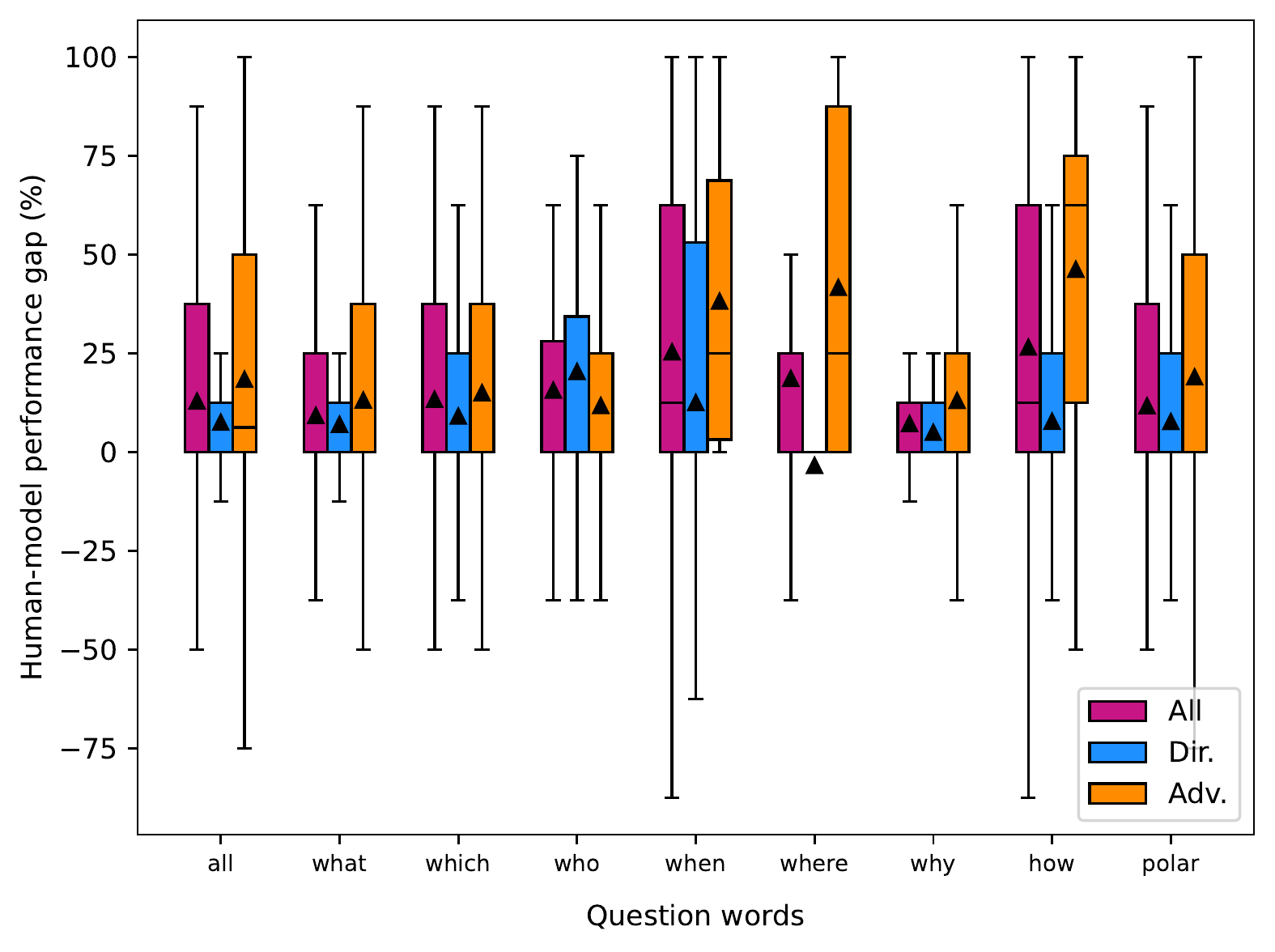}
    \caption{
        Question words and human--model performance gap.
        The triangle markers indicate mean values and the black bars indicate medians.
    }
    \label{fig:qtype-gap-boxplot}
\end{figure}

\begin{figure}[t]
    \includegraphics[width=\linewidth]{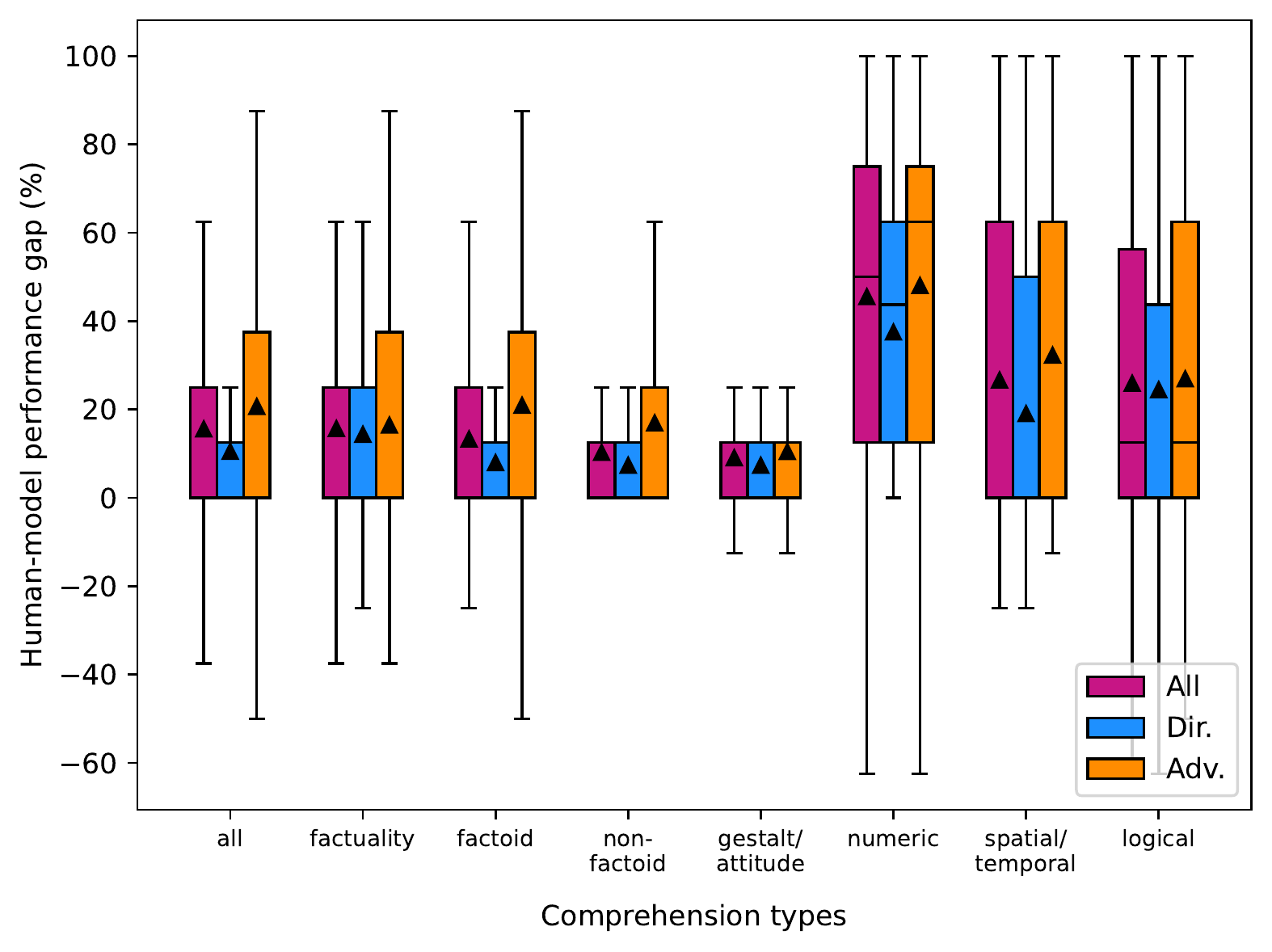}
    \caption{
        Comprehension types and human--model performance gap.
        The triangle markers indicate mean values and the black bars indicate medians.
    }
    \label{fig:reasoning-gap-boxplot}
\end{figure}

\begin{figure*}[tbh]
    \centering
    \includegraphics[width=1.0\linewidth]{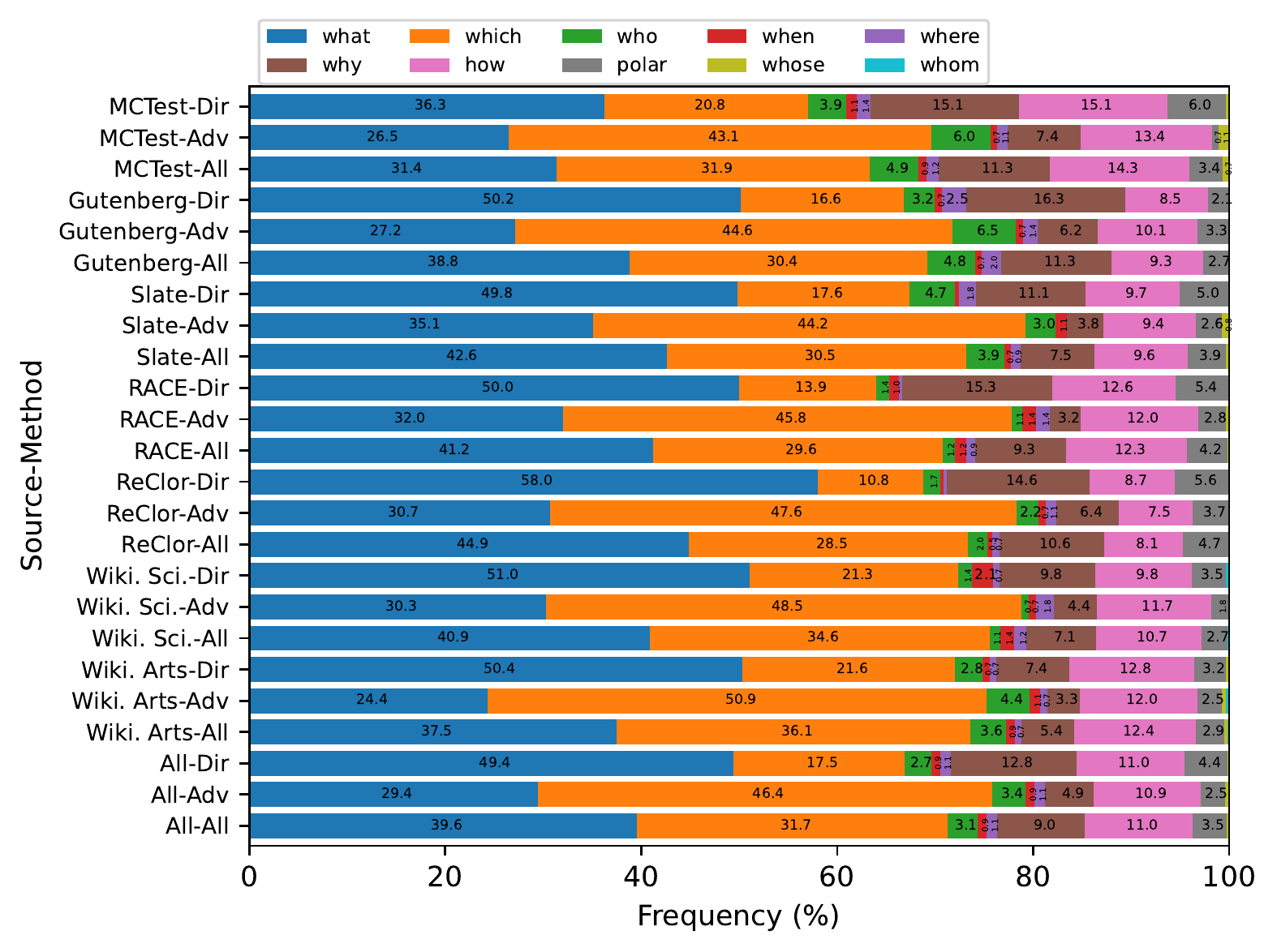}
    \caption{
        Frequencies of question words ($wh$-words) across passage sources and collection methods.
    }
    \label{fig:question-type-detail}
\end{figure*}

Figure~\ref{fig:qtype-pie-methods} shows the frequency of the question words and the two subsequent words for each collection method.
Figures~\ref{fig:qtype-gap-boxplot} and \ref{fig:reasoning-gap-boxplot} show the box plots between human--model performance gap and questions words or comprehension types, respectively.
Figures~\ref{fig:question-type-detail} and \ref{fig:reasoning-type-detail} show the frequency of question words and comprehension types, respectively, across the passage sources and collection methods.
In the comprehension types annotation, a question can have multiple labels.
Therefore, the sum of the frequencies may exceed 100\%.

The definitions of the comprehension types are as follows:
\begin{enumerate}
    \item \textbf{Factuality} (\textit{true/false/likely}) is reasoning of which answer option most (or least) describes facts or events in a given passage. 
    \item \textbf{Factoid} simply asks about described events or entities, typically with typical \textit{what} questions. 
    \item \textbf{Non-factoid} is related to \textit{why} and \textit{how} questions, such as ones asking about causality, a character's attitude, or the process of described events. 
    \item \textbf{Gestalt/Attitude} asks about the summary, theme, or conclusion of the content of a given passage or the author's attitude towards it. 
    \item \textbf{Numeric} indicates questions that require arithmetic reasoning. 
    \item \textbf{Spatial/Temporal} is related to the understanding of places and locations (spatial) or the temporal order or duration (temporal) of described events. 
    \item \textbf{Logical} is pertinent to logical reasoning and arguments described in a passage. 
\end{enumerate}

\section{Human Accuracy as Question Difficulty}
\label{app:human_difficulty}

\begin{table}[t]
    \begin{tabular}{lcc} \toprule
    Aspects & $r$ & $p$ \\ \midrule
    Passage length & 0.009 & 0.59 \\
    Flesch--Kincaid grade & -0.06 & $<$0.001 \\
    Elapsed time for answering & -0.16 & $<$0.001 \\
    Elapsed time for writing & -0.04 & 0.007 \\
    Syntactic surprisal & -0.01 & 0.53 \\
    Semantic surprisal & -0.001 & 0.93 \\
    Average word frequency & 0.004 & 0.82 \\
    \bottomrule
    \end{tabular}
    \caption{Pearson's correlation $r$ and its $p$-value between the human accuracy and textual aspects.}
    \label{fig:human-text-aspects}
\end{table}

\begin{figure}[t]
    \includegraphics[width=\linewidth]{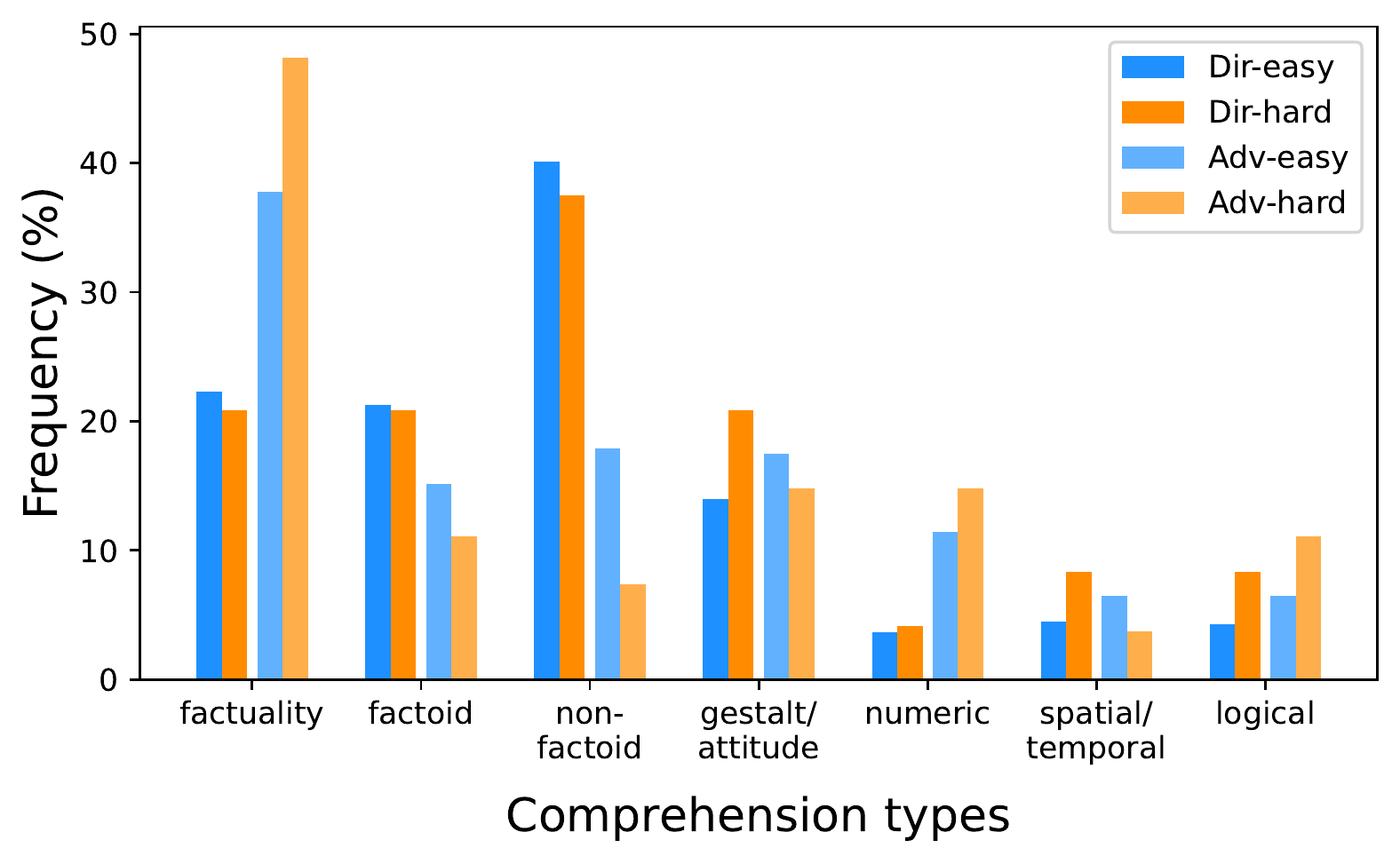}
    \caption{
        Frequency of comprehension types in easy and hard examples as determined by the question difficulty for humans for each collection method.
    }
    \label{fig:human-difficulty-to-reasoning}
\end{figure}

We compute a similar linguistic analysis using the average human accuracy as the difficulty of the questions.
Table~\ref{fig:human-text-aspects} shows Pearson's correlation $r$ and its $p$-value between the human accuracy (as the question difficulty) and textual aspects.
Just as when using the human--model gap, we do not observe any strong correlations except for the elapsed time for answering that shows a weak negative correlation, which means difficult-for-human questions take slightly longer for answering.
Figure~\ref{fig:human-difficulty-to-reasoning} shows the frequency of comprehension types in easy and hard examples with regard to the question difficulty for humans.

\section{Examples of Collected Questions}
\label{app:examples}

\begin{table*}[t]
    \def\arraystretch{1.5}
    \begin{tabular}{p{0.21\linewidth}p{0.72\linewidth}}\toprule
        Comprehension Type \newline (source, difficulty) & Example \\ \midrule
        Factuality \newline (Gutenberg, easy) & Q: Which of the following is not mentioned in the passage? \newline A: \checkbox An Earl lived in a house that had a relatively low profile. / \checkbox There were some other buildings near the Manor. / \checkbox Scroope is a village that is closely linked to an Earl's home. / \checkedbox Scroope Manor was sold to the village by the Earl.  \\
        Factoid \newline (Wiki. science, easy) & Q: What helps many fish keep their buoyancy in water? \newline A: \checkbox muscles on either side of the backbone / \checkbox fins / \checkedbox a swim bladder / \checkbox a streamlined body \\ %
        Non-factoid \newline (Wiki. arts, hard) & Q: How did a major portion of English words enter the English language? \newline A: \checkbox French speakers can understand many English words without having to undergo any orthographical change. / \checkbox Many words in Old English are from Old Norse. / \checkedbox About one-third of words in English entered the language from the long contact between French and English. / \checkbox Romance languages have "Latinate" roots. \\
        Gestalt/Attitude \newline (Slate, easy) & Q: Which of the following is a criticism the author has about Dick Riordan? \newline A: \checkbox He's not transparent about his typical lunch looks like, which highlights his lack of wisdom. / \checkedbox He's okay syphoning resources from elsewhere to himself for personal gain. / \checkbox Much like Hillary Clinton, he lacks any sort of coherent persona. / \checkbox He is responsible for the vast swaths of one-story buildings that cover the entire landscape of L.A. \\
        Numeric \newline (RACE, hard) & Q: How old was Mary Shelley when she died? \newline A: \checkbox Mary Shelley was in her thirties when she died. / \checkbox Mary Shelley died when she was forty four years old. / \checkedbox Mary Shelley died when she was in her fifties. / \checkbox Mary Shelley lived well into her eighties before she died. \\
        Spatial/Temporal \newline (MCTest, easy) & Q: When did it start to rain? \newline A: \checkedbox It started to rain after Will ate his biscuit and jam. / \checkbox It started to rain after Will heard the thunder. / \checkbox It started to rain while Will was at the store. / \checkbox It started to rain on Will's walk home from the store. \\
        Logical \newline (ReClor, hard) & Q: Which statement, if true, would weaken the conclusion of the passage? \newline A: \checkbox Archaeologists have found remains of shipwrecks from 2000 BC between Crete and southern Greece. / \checkedbox The earliest bronze artifacts found in southern Greece date to 3000 BC. / \checkbox The Minoans were far more accomplished in producing bronzeware than any other civilization in the area at the time. / \checkbox The capacity of Minoan bronze furnaces was extraordinarily large compared to other societies in 2000 BC. \\
        \bottomrule
    \end{tabular}
    \caption{Examples of each comprehension type taken from our collected data.}
    \label{tab:examples}
\end{table*}

Table~\ref{tab:examples} shows examples of questions and options for each comprehension type.
After extracting the question words, we review about 100 questions to collect keywords that determine comprehension type (e.g., ``reason'' for \textit{non-factoid},``best summarize'' for \textit{gestalt/attitude} and ``if'' for \textit{logical}).
We then write simple rules that highlight these keywords, which help us manually annotate the remaining questions within approximately five hours.

\section{Writing Instructions and Examples}
\label{app:instructions}

\begin{figure*}[t]
    \centering
    \frame{\includegraphics[width=1.0\linewidth]{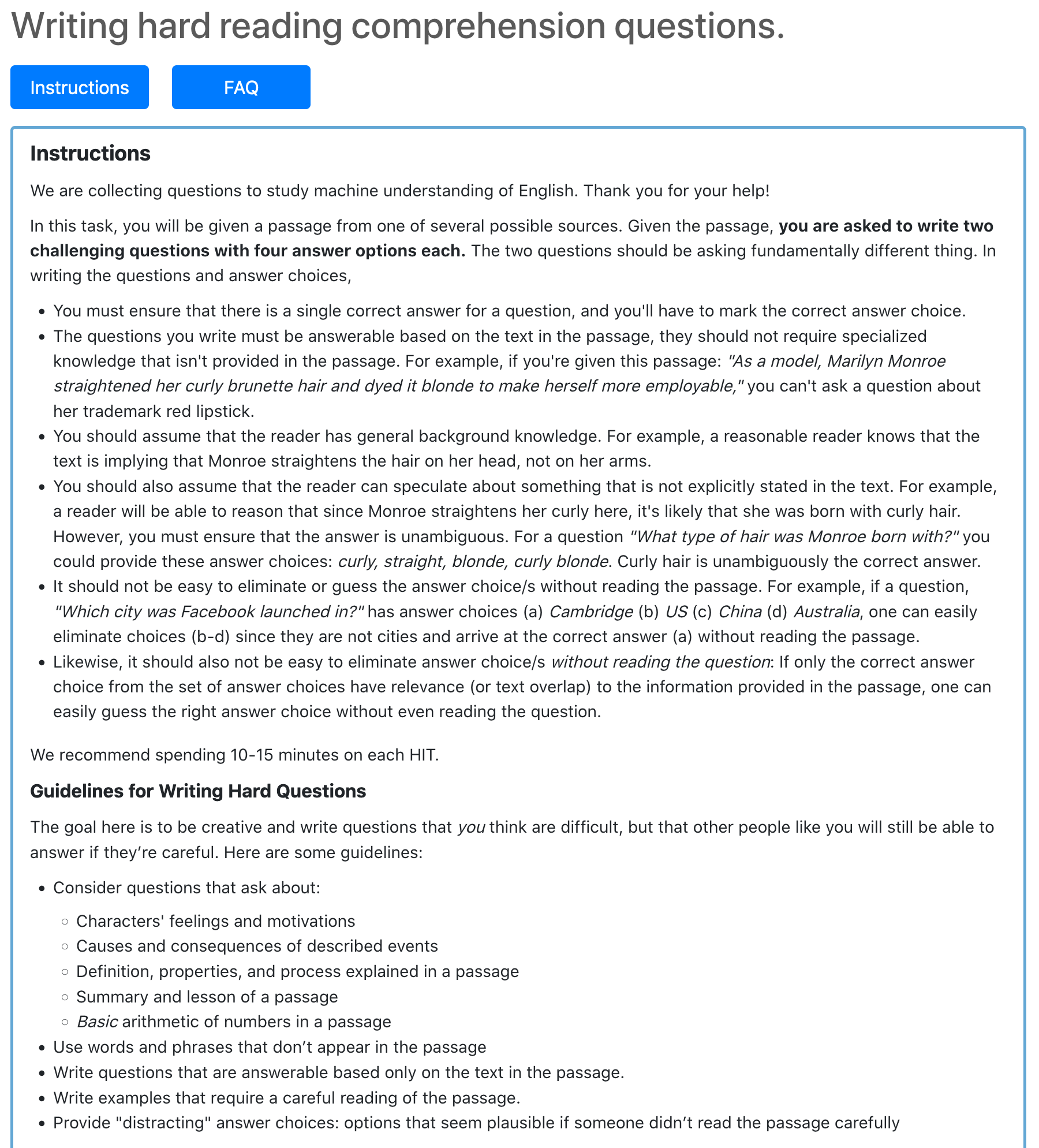}}
    \caption{
        Instructions of the writing task.
    }
    \label{fig:instructions}
\end{figure*}

\begin{figure*}[t]
    \centering
    \frame{\includegraphics[width=1.0\linewidth]{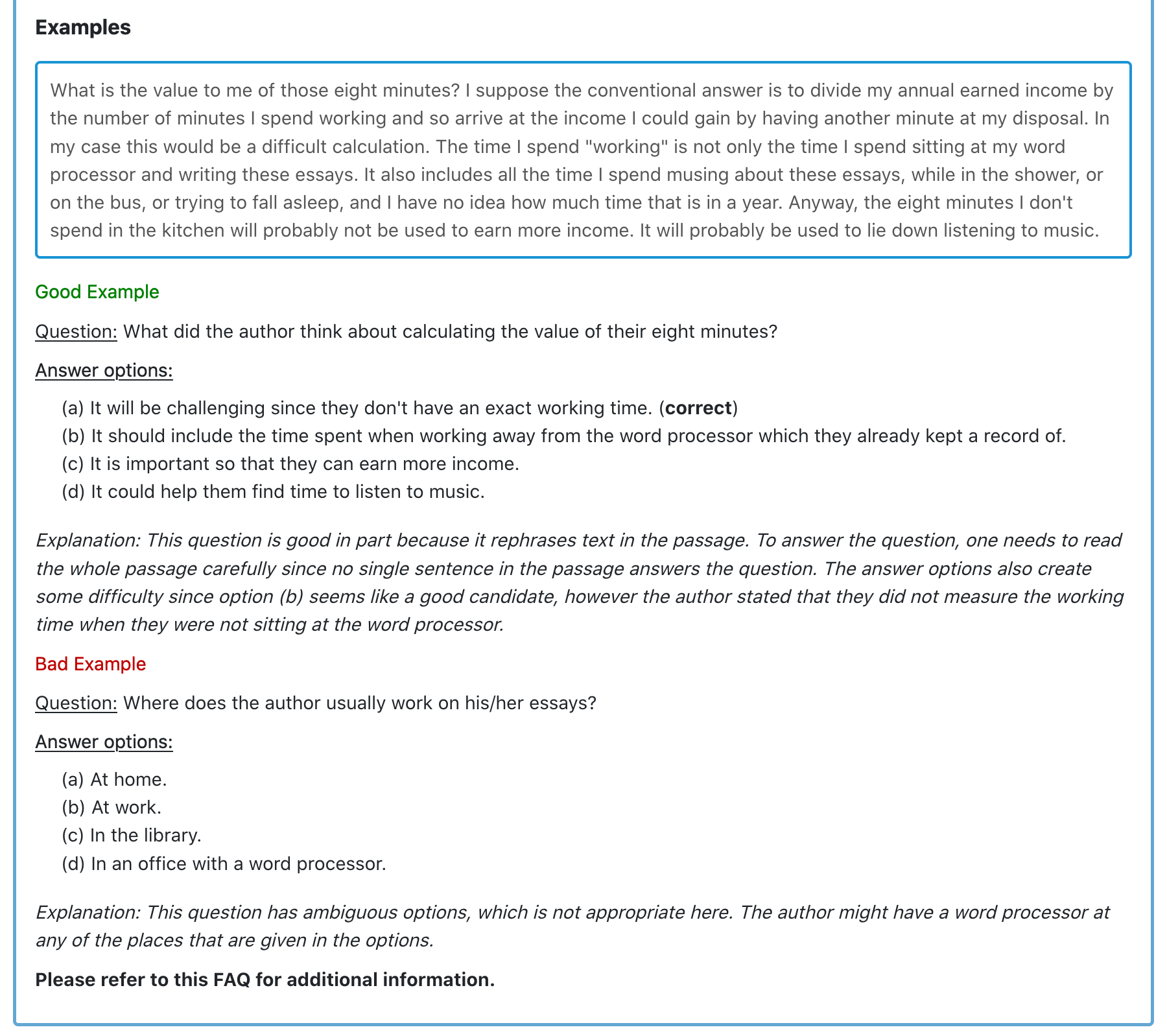}}
    \caption{
        Good and bad examples included in the instructions of the writing task.
    }
    \label{fig:examples}
\end{figure*}

\begin{figure*}[t]
    \centering
    \frame{\includegraphics[width=1.0\linewidth]{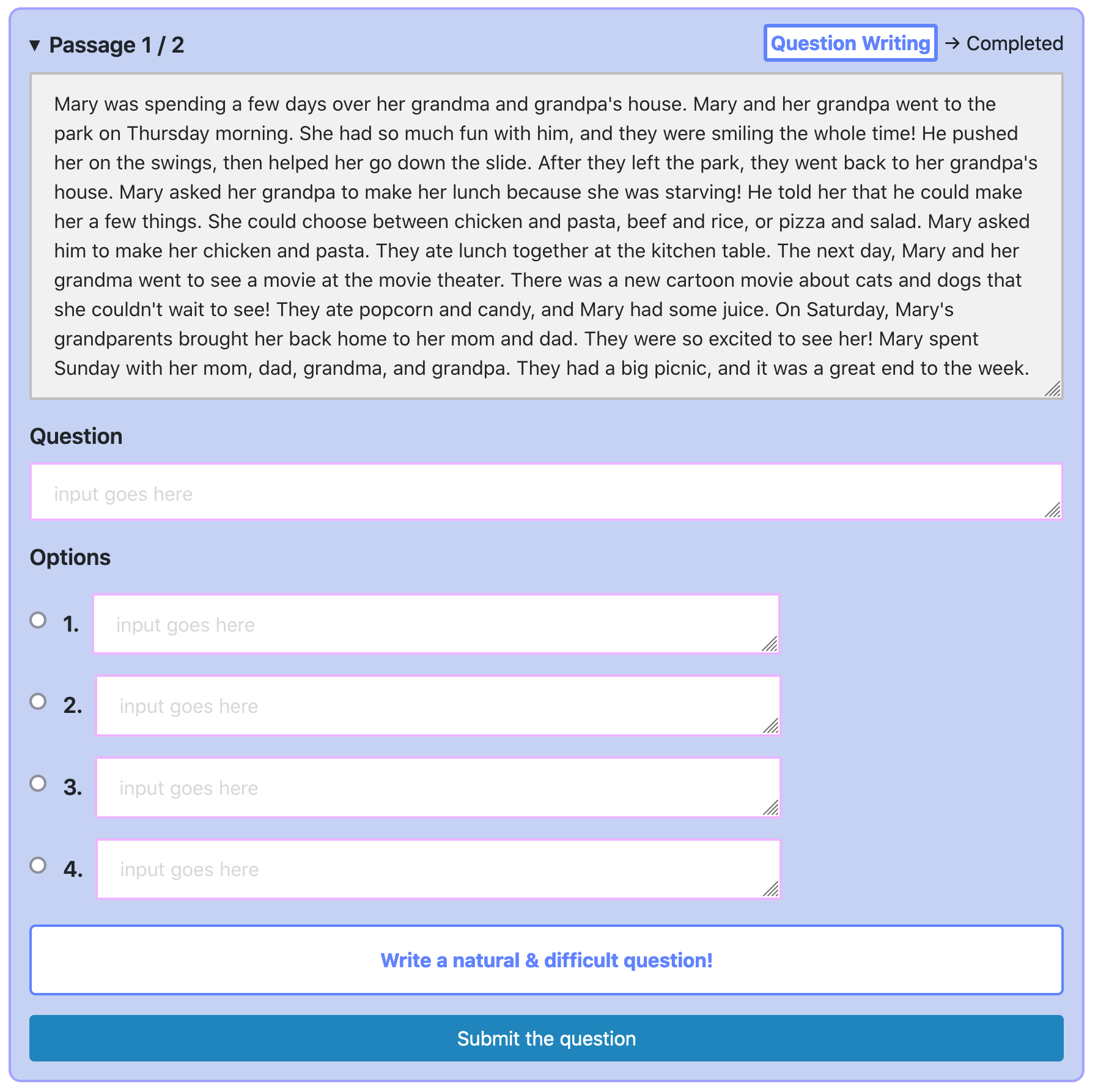}}
    \caption{
        Interface of the writing task.
    }
    \label{fig:interface}
\end{figure*}

Figures~\ref{fig:instructions},~\ref{fig:examples}, and~\ref{fig:interface} show the instructions, good and bad examples, and task interface provided to the crowdworkers in our data collection.

\end{document}

%% file: main_result.tex
     &  & \multicolumn{5}{c}{All valid examples} & \multicolumn{5}{c}{High-agreement portion} \\ \cmidrule(lr){3-7} \cmidrule(lr){8-12}
\multicolumn{2}{c}{Source \hphantom{\&} Method} & Human       & UniQA & DeBERTa & M-Avg.       & $\Delta$      & Human        & UniQA & DeBERTa & M-Avg.       & $\Delta$   \\ \midrule
MCTest & Dir. & 89.1 & 68.3 & 84.5 & \underline{78.1} & 11.0 & 95.0 & 71.5 & 88.2 & \underline{81.5} & \textbf{13.5} \\
\rowcolor{gray!15}
\hphantom{MCTest} & Adv. & \textbf{93.6} & 26.5 & 75.3 & 66.6 & \textbf{27.1} & \textbf{96.5} & 27.9 & 78.6 & 68.2 & \textbf{28.3} \\
\rowcolor{gray!30}
\hphantom{MCTest} & Total & \textbf{91.4} & 47.4 & 79.9 & 72.3 & \textbf{19.0} & \textbf{95.8} & 49.3 & 83.3 & \underline{74.7} & \textbf{21.1} \\
Gutenberg & Dir. & 85.2 & 70.7 & 84.5 & 79.9 & 5.3 & 92.8 & 75.0 & 88.5 & 83.4 & 9.4 \\
\rowcolor{gray!15}
\hphantom{Gutenberg} & Adv. & 83.0 & 26.4 & 80.1 & \textbf{69.7} & 13.3 & \underline{87.5} & 28.3 & 82.6 & 72.9 & \underline{14.6} \\
\rowcolor{gray!30}
\hphantom{Gutenberg} & Total & 84.1 & 48.8 & 82.3 & 74.8 & 9.3 & \underline{90.3} & 53.1 & 85.7 & 78.4 & 11.9 \\
Slate & Dir. & \underline{84.9} & 72.4 & 88.9 & 84.1 & \underline{0.8} & \underline{90.7} & 74.6 & 91.7 & 87.0 & 3.8 \\
\rowcolor{gray!15}
\hphantom{Slate} & Adv. & \underline{82.6} & 26.0 & 71.7 & 69.4 & \underline{13.2} & 92.9 & 27.9 & 76.0 & \textbf{73.8} & 19.1 \\
\rowcolor{gray!30}
\hphantom{Slate} & Total & \underline{83.8} & 49.8 & 80.5 & \textbf{77.0} & \underline{6.8} & 91.8 & 52.6 & 84.3 & \textbf{80.8} & \underline{11.0} \\
RACE & Dir. & 91.2 & 70.4 & 85.0 & 80.8 & 10.3 & 95.4 & 74.8 & 90.4 & 84.6 & 10.8 \\
\rowcolor{gray!15}
\hphantom{RACE} & Adv. & 89.4 & 28.9 & 69.4 & \underline{65.0} & 24.4 & 94.3 & 31.0 & 73.8 & \underline{67.3} & 27.0 \\
\rowcolor{gray!30}
\hphantom{RACE} & Total & 90.3 & 50.0 & 77.3 & 73.1 & 17.3 & 94.9 & 53.3 & 82.2 & 76.1 & 18.8 \\
ReClor & Dir. & \textbf{94.1} & 72.6 & 88.5 & 80.6 & \textbf{13.5} & \textbf{96.9} & 79.6 & 91.1 & 84.4 & 12.5 \\
\rowcolor{gray!15}
\hphantom{ReClor} & Adv. & 83.9 & 29.2 & 71.5 & 66.3 & 17.6 & 88.8 & 32.4 & 74.5 & 71.3 & 17.5 \\
\rowcolor{gray!30}
\hphantom{ReClor} & Total & 89.2 & 51.7 & 80.4 & 73.7 & 15.5 & 93.2 & 58.1 & 83.5 & 78.5 & 14.8 \\
Wiki. Sci. & Dir. & 90.6 & 75.9 & 90.6 & 83.2 & 7.3 & 95.8 & 79.0 & 94.9 & 87.3 & 8.5 \\
\rowcolor{gray!15}
\hphantom{Wiki. Sci.} & Adv. & 84.3 & 27.4 & 75.2 & 65.6 & 18.8 & 92.8 & 29.4 & 77.2 & 68.3 & 24.5 \\
\rowcolor{gray!30}
\hphantom{Wiki. Sci.} & Total & 87.5 & 52.1 & 83.0 & 74.6 & 12.9 & 94.4 & 56.3 & 86.8 & 78.6 & 15.8 \\
Wiki. Arts & Dir. & 88.3 & 76.2 & 88.7 & \textbf{84.2} & 4.1 & 91.5 & 77.0 & 92.5 & \textbf{88.1} & \underline{3.4} \\
\rowcolor{gray!15}
\hphantom{Wiki. Arts} & Adv. & 83.3 & 25.5 & 73.8 & 69.4 & 13.9 & 91.4 & 25.8 & 75.8 & 71.7 & 19.7 \\
\rowcolor{gray!30}
\hphantom{Wiki. Arts} & Total & 85.8 & 51.2 & 81.3 & 76.9 & 8.9 & 91.5 & 52.3 & 84.5 & 80.2 & 11.2 \\ \midrule
All sources & Dir. & 89.0 & 72.4 & 87.2 & 81.6 & 7.5 & 94.0 & 75.9 & 91.0 & 85.2 & 8.8 \\
\rowcolor{gray!15}
\hphantom{All sources} & Adv. & 85.7 & 27.1 & 73.8 & 67.4 & 18.3 & 92.0 & 29.0 & 76.9 & 70.5 & 21.5 \\
\rowcolor{gray!30}
\hphantom{All sources} & Total & 87.4 & 50.2 & 80.7 & 74.6 & 12.8 & 93.1 & 53.6 & 84.3 & 78.2 & 14.9 \\

%% file: dataset_stats.tex
Source & Method & Valid  & High  \\ \midrule
MCTest & Dir. & 91.6 & 71.3 \\
\rowcolor{gray!15}
\hphantom{MCTest} & Adv. & 91.3 & 73.9 \\
\rowcolor{gray!30}
\hphantom{MCTest} & Total & 91.5 & 72.6 \\
Gutenberg & Dir. & 91.3 & 67.1\\
\rowcolor{gray!15}
\hphantom{Gutenberg} & Adv. & 89.0 & 59.4  \\
\rowcolor{gray!30}
\hphantom{Gutenberg} & Total & 90.2 & 63.2 \\
Slate & Dir. & 90.0 & 66.1 \\
\rowcolor{gray!15}
\hphantom{Slate} & Adv. & 85.5 & 59.0 \\
\rowcolor{gray!30}
\hphantom{Slate} & Total & 87.7 & 62.6 \\
RACE & Dir. & 94.8 & 70.3 \\
\rowcolor{gray!15}
\hphantom{RACE} & Adv. & 91.6 & 67.7 \\
\rowcolor{gray!30}
\hphantom{RACE} & Total & 93.2 & 69.0 \\
ReClor & Dir. & 92.9 & 72.6 \\
\rowcolor{gray!15}
\hphantom{ReClor} & Adv. & 86.1 & 60.6 \\
\rowcolor{gray!30}
\hphantom{ReClor} & Total & 89.5 & 66.6 \\
Wiki. Sci. & Dir. & 92.3 & 69.0  \\
\rowcolor{gray!15}
\hphantom{Wiki. Sci.} & Adv. & 88.4 & 58.1 \\
\rowcolor{gray!30}
\hphantom{Wiki. Sci.} & Total & 90.3 & 63.5 \\
Wiki. Arts & Dir. & 91.0 & 64.5 \\
\rowcolor{gray!15}
\hphantom{Wiki. Arts} & Adv. & 88.7 & 60.0  \\
\rowcolor{gray!30}
\hphantom{Wiki. Arts} & Total & 89.8 & 62.3  \\ \midrule
All sources & Dir. & 92.0 & 68.7 \\
\rowcolor{gray!15}
\hphantom{All sources} & Adv. & 88.7 & 62.7 \\
\rowcolor{gray!30}
\hphantom{All sources} & Total & 90.3 & 65.7 \\

%% file: transfer_results.tex
Source & Method & Valid & High \\ \midrule

MCTest & Dir. & 70.7$_{+6.9}$ & 72.2$_{+6.6}$ \\
\rowcolor{gray!15}
\hphantom{MCTest} & Adv. & 65.6$_{+1.8}$ & 68.0$_{+2.5}$ \\
Gutenberg & Dir. & 79.2$_{+5.6}$ & 82.1$_{+5.5}$ \\
\rowcolor{gray!15}
\hphantom{Gutenberg} & Adv. & 76.0$_{+2.4}$ & 79.6$_{+3.0}$ \\
Slate & Dir. & 77.1$_{+3.8}$ & 79.1$_{+3.1}$ \\
\rowcolor{gray!15}
\hphantom{Slate} & Adv. & 74.2$_{+0.8}$ & 77.0$_{+1.0}$ \\
RACE & Dir. & 78.2$_{+8.6}$ & 79.6$_{+9.3}$ \\
\rowcolor{gray!15}
\hphantom{RACE} & Adv. & 71.8$_{+2.3}$ & 72.6$_{+2.2}$ \\
ReClor & Dir. & 74.6$_{+1.6}$ & 76.1$_{+1.0}$ \\
\rowcolor{gray!15}
\hphantom{ReClor} & Adv. & 72.6$_{-0.4}$ & 74.6$_{-0.5}$ \\
Wiki. Sci. & Dir. & 78.5$_{+7.7}$ & 79.4$_{+8.5}$ \\
\rowcolor{gray!15}
\hphantom{Wiki. Sci.} & Adv. & 74.8$_{+4.1}$ & 74.9$_{+4.0}$ \\
Wiki. Arts & Dir. & 80.7$_{+6.6}$ & 79.7$_{+5.4}$ \\
\rowcolor{gray!15}
\hphantom{Wiki. Arts} & Adv. & 75.3$_{+1.2}$ & 75.2$_{+1.0}$ \\

%% file: partial_input.tex
Source & Meth. & P+A & Q+A & A only \\ \midrule
MCTest & Dir. & 73.3$_{-14.9}$ & 39.8$_{-48.4}$ & 29.4$_{-58.8}$ \\
\rowcolor{gray!15}
\hphantom{MCTest} & Adv. & 55.5$_{-23.1}$ & 41.5$_{-37.1}$ & 34.5$_{-44.1}$ \\
\rowcolor{gray!30}
\hphantom{MCTest} & Total & 64.2$_{-19.1}$ & 40.7$_{-42.7}$ & 32.0$_{-51.3}$ \\
Gutenberg & Dir. & 75.5$_{-13.0}$ & 40.9$_{-47.6}$ & 31.7$_{-56.7}$ \\
\rowcolor{gray!15}
\hphantom{Gutenberg} & Adv. & 55.4$_{-27.2}$ & 42.4$_{-40.2}$ & 34.2$_{-48.4}$ \\
\rowcolor{gray!30}
\hphantom{Gutenberg} & Total & 66.1$_{-19.6}$ & 41.6$_{-44.1}$ & 32.9$_{-52.8}$ \\
Slate & Dir. & 72.7$_{-19.0}$ & 45.9$_{-45.9}$ & 32.7$_{-59.0}$ \\
\rowcolor{gray!15}
\hphantom{Slate} & Adv. & 54.1$_{-21.9}$ & 44.3$_{-31.7}$ & 33.9$_{-42.1}$ \\
\rowcolor{gray!30}
\hphantom{Slate} & Total & 63.9$_{-20.4}$ & 45.1$_{-39.2}$ & 33.2$_{-51.0}$ \\
RACE & Dir. & 75.7$_{-14.7}$ & 49.5$_{-40.8}$ & 36.2$_{-54.1}$ \\
\rowcolor{gray!15}
\hphantom{RACE} & Adv. & 49.0$_{-24.8}$ & 43.3$_{-30.5}$ & 31.9$_{-41.9}$ \\
\rowcolor{gray!30}
\hphantom{RACE} & Total & 62.6$_{-19.6}$ & 46.5$_{-35.7}$ & 34.1$_{-48.1}$ \\
ReClor & Dir. & 78.7$_{-12.4}$ & 44.4$_{-46.7}$ & 35.1$_{-56.0}$ \\
\rowcolor{gray!15}
\hphantom{ReClor} & Adv. & 55.9$_{-18.6}$ & 41.5$_{-33.0}$ & 26.6$_{-47.9}$ \\
\rowcolor{gray!30}
\hphantom{ReClor} & Total & 68.3$_{-15.3}$ & 43.1$_{-40.4}$ & 31.2$_{-52.3}$ \\
Wiki. Sci. & Dir. & 76.2$_{-18.7}$ & 45.8$_{-49.1}$ & 33.2$_{-61.7}$ \\
\rowcolor{gray!15}
\hphantom{Wiki. Sci.} & Adv. & 54.4$_{-22.8}$ & 35.6$_{-41.7}$ & 26.7$_{-50.6}$ \\
\rowcolor{gray!30}
\hphantom{Wiki. Sci.} & Total & 66.2$_{-20.6}$ & 41.1$_{-45.7}$ & 30.2$_{-56.6}$ \\
Wiki. Arts & Dir. & 70.0$_{-22.5}$ & 49.0$_{-43.5}$ & 44.5$_{-48.0}$ \\
\rowcolor{gray!15}
\hphantom{Wiki. Arts} & Adv. & 53.8$_{-22.0}$ & 44.6$_{-31.2}$ & 26.3$_{-49.5}$ \\
\rowcolor{gray!30}
\hphantom{Wiki. Arts} & Total & 62.2$_{-22.3}$ & 46.9$_{-37.6}$ & 35.8$_{-48.7}$ \\ \midrule
All src. & Dir. & 74.6$_{-16.5}$ & 45.0$_{-46.0}$ & 34.7$_{-56.3}$ \\
\rowcolor{gray!15}
\hphantom{All src.} & Adv. & 54.0$_{-22.9}$ & 41.9$_{-35.0}$ & 30.6$_{-46.3}$ \\
\rowcolor{gray!30}
\hphantom{All src.} & Total & 64.8$_{-19.5}$ & 43.6$_{-40.8}$ & 32.8$_{-51.6}$ \\